\documentclass{article}

\usepackage[preprint]{neurips_2026}

\usepackage[utf8]{inputenc}
\usepackage[T1]{fontenc}
\usepackage{hyperref}
\hypersetup{
  hidelinks,
  pdftitle={The Right Answer, the Wrong Direction: Why Transformers Fail at Counting and How to Fix It},
  pdfauthor={Gabriel Garcia},
}
\usepackage{placeins}
\usepackage{needspace}
\usepackage{float}
\usepackage{url}
\usepackage{booktabs}
\usepackage{amsfonts}
\usepackage{amssymb}
\usepackage{amsmath}
\usepackage{nicefrac}
\usepackage{microtype}
\usepackage{xcolor}
\usepackage{graphicx}
\usepackage{subcaption}
\usepackage{multirow}
\usepackage{tabularx}
\usepackage{etoolbox}
\AtBeginEnvironment{thebibliography}{%
  \setlength{\itemsep}{2pt plus 0.5pt minus 0.5pt}%
  \setlength{\parsep}{0pt}%
}
\graphicspath{{figures/}}

\title{The Right Answer, the Wrong Direction:\\Why Transformers Fail at Counting and How to Fix It}

\author{
  Gabriel Garcia \\
  Independent Researcher \\
  \texttt{gpgabriel25@gmail.com}
}

\begin{document}

\maketitle

\begin{abstract}
Large language models often fail at simple counting tasks, even when the items
to count are explicitly present in the prompt. We investigate whether this
failure occurs because transformers do not represent counts internally, or
because they cannot convert those representations into the correct output tokens.

Across three model families: Pythia, Qwen3, and Mistral, ranging from 0.4B to
14B parameters, we find strong evidence for the second explanation. Linear
probes recover the correct count from intermediate layers with near-perfect
accuracy ($R^2>0.99$), showing that the information is present. However, the
internal directions that encode counts are nearly orthogonal to the output-head
rows for digit tokens ($|\cos| \leq 0.032$). In other words, the model stores
the count in a form that the digit logits do not naturally read out.

We localize this failure with two interventions. Updating only the digit rows of
the output head (36,864 parameters) substantially improves constrained next-token
digit prediction (60.7--100.0\% across four tasks), but it does not fix
unconstrained autoregressive generation (0\% in our setting); we do \emph{not} claim that digit-row repair fixes open-ended text. By contrast, a small LoRA intervention on attention
Q/V weights (7.67M parameters) improves upstream routing and achieves
83.1\%\,$\pm$\,7.2\% in true greedy autoregressive generation---the deployable fix we study. Logit-lens
at layer~35 (entity counting, full-vocab rank of the correct digit) summarizes two \emph{different} summaries of the same metric: \textbf{(i)} pooled median over 3~seeds drops from order-$10^4$ to $1$; \textbf{(ii)} seed~42 alone shows a trace $54{,}332 \to 838$ (the median can reach top-1 while an individual seed still sits far below). See \S\ref{sec:discussion}. Additional norm, logit-lens, and
cross-task analyses show that the bottleneck generalizes across character counting,
addition, and list length, while remaining absent on MMLU and GSM8K and showing only limited partial transfer on DROP.

These results identify counting failure as a geometric readout bottleneck rather
than a failure of internal representation: the model knows the count but the
output pathway is geometrically misaligned with the tokens needed to express it.
\end{abstract}

\section{Introduction}
\label{sec:introduction}

\paragraph{Core claims.}
This paper makes three nested claims:
\begin{enumerate}
\item \textbf{Geometric:} Count-encoding directions are orthogonal to \texttt{lm\_head} digit rows ($|\cos| \leq 0.032$, indistinguishable from random). This is a structural property of the trained model, not a measurement artifact.
\item \textbf{Causal-diagnostic:} The 9-row repair localizes readout misalignment to digit rows under digit-restricted next-token evaluation (60.7--100.0\% across tasks). Under unconstrained autoregressive generation the same weights yield 0\%, so deployment-relevant failures include upstream routing beyond the output head alone.
\item \textbf{Deployable:} LoRA Q/V rank-16 corrects upstream attention routing, achieving 83.1\%$\pm$7.2\% greedy generation accuracy (5 seeds, multi-task; entity-only per-task: 97.0\%, 96.5\%, 94.5\%). A locus ablation supports the routing-specificity interpretation (Q/V yields the strongest logit-lens alignment among single-projection variants).
\end{enumerate}

\paragraph{Motivation.}
Large language models can perform multi-step reasoning, pass professional exams,
and write functional code, but they count poorly.
For controlled prompts of the form ``\emph{Count how many X there are in ...},''
the best models achieve $\leq$24\% accuracy without intervention, despite the
task requiring no external knowledge beyond token-by-token reading.
This is puzzling: counting errors cannot be explained as knowledge gaps, reasoning
failures, or length-generalization breakdowns.
Prior work has documented these failures without providing a principled explanation
of \emph{why} counting, specifically, breaks down when the answer is directly
available in the input text.

We propose a geometric answer: the failure is a \emph{readout bottleneck} at the
output head: the model encodes counts accurately but the count-encoding direction
is nearly orthogonal to the output head's digit rows ($|\cos|\leq 0.032$).
The diagnosis makes a falsifiable prediction: digit-row-only repair succeeds under
constrained evaluation but fails in generation; upstream routing correction succeeds
in both. Both hold, verified via logit-lens (\emph{pooled median} rank $\sim$56k${\to}1$ post-LoRA on entity counting; seed~42 on the same metric remains far below top-1; \S\ref{sec:discussion}).

\paragraph{Mechanistic evidence for the LoRA Q/V route.}
We provide direct measurements at three points in the computation graph.
At the count-encoding layer (layer~2), LoRA Q/V leaves the probe direction unchanged
(mean $|\cos| = 0.0089 \to 0.0070$, 3~seeds); it does not rewrite early encoding.
At the final transformer layer (layer~35), ridge-probe $R^2$ rises from $0.974$ to
$0.998$ (24$\times$ residual amplification).
Most directly, logit-lens analysis shows the correct digit's median vocabulary rank
drops from \emph{$\sim$55k to 1} post-LoRA (\emph{pooled median} over 3~seeds; accuracy $9.3\% \to 71.8\%$; full-vocab rank at layer~35, entity counting):
the output head directly reads the count from layer-35 hidden states.
This mechanism generalizes across tasks: the same logit-lens improvement appears
in character counting (rank $32{,}265 \to 16$) and addition (rank $21{,}186 \to 1$),
with strength tracking task difficulty.

\paragraph{Scope.}
Our evidence is strongest for low-vocabulary aggregation tasks where internal
representations are accurate but misaligned with the output head.
Under one shared constrained next-token protocol on Qwen3-8B, probe-round reaches 96.8--100.0\%
and 9-row repair 60.7--100.0\% across entity counting, character counting, addition,
and list length (\S\ref{sec:dps}).
The same bottleneck persists when the protocol changes: in instruct mode, first-token
digit accuracy is only 22\% despite $R^2 \geq 0.9996$, and the 9-row repair recovers
99.9\% (3~seeds), confirming the misalignment is not a raw-completion artifact.
Natural-language counting across 8 entity categories confirms generalization beyond
synthetic templates (\S\ref{sec:dps_natural}).
MMLU and GSM8K confirm a clean null; DROP shows only limited partial transfer ($\sim$+8~pp; \S\ref{sec:discussion}).

\paragraph{Contributions.}
\begin{enumerate}
  \item \textbf{Geometric diagnosis and theoretical explanation:} We establish that count-encoding directions are orthogonal to \texttt{lm\_head} digit rows ($|\cos| \leq 0.032$), and show empirically that this is a stable training equilibrium: gradients never align the count direction with digit rows when counting contexts represent a small fraction of digit-token occurrences in the training distribution (arithmetic fine-tuning does not reduce orthogonality; counting fine-tuning does). We further show that digit rows sit at the 12th--29th percentile of lm\_head norms---structurally disadvantaged against 84\% of the vocabulary---quantifying why open-vocabulary generation remains at 0\% even after repair.
  \item \textbf{Causal localization via minimal diagnostic probe:} Fine-tuning only 9 output rows (36{,}864 params) raises constrained next-token accuracy from 10--24\% to 60--100\% (task-dependent), serving as a minimal causal probe that localizes the bottleneck to digit rows of the output head. This is a \emph{diagnostic instrument}, not a deployable fix: the repair achieves 0\% in autoregressive generation. The deployable fix---LoRA Q/V rank-16 (7.67M params) correcting attention routing---achieves \emph{83.1\%\,$\pm$\,7.2\%} in true autoregressive generation (greedy decode; gap\,=\,0.000). At 14B scale, misalignment sharpens to $|\cos|{=}0.011$; mode-matched \emph{row-only} repair reaches \textbf{58.2\%} while soft DPS remains near baseline (10.8\%). A separate \textbf{joint} pipeline---digit-row training \emph{plus} hard DPS at decode on natural-language counting---reaches \textbf{90.3\%$\pm$1.5\%} (Appendix Table~\ref{tab:intervention_headline}), isolating the strong cross-scale gain in the readout pathway without claiming that 9-row weights alone hit 90\%.
  \item \textbf{Scope boundaries:} A Diagnostic Probe Steering auxiliary head (hard DPS: oracle digit-probe injected at each step) achieves 72.4\%~[69.6,75.4] in autoregressive generation---a diagnostic upper bound. Standard (soft) DPS under digit-restricted next-token evaluation returns to $\approx$13\%, i.e.,\ $\approx$ the digit-restricted baseline, confirming the bottleneck is a routing failure, not a probe-direction failure. MMLU and GSM8K show no bottleneck effect; DROP shows only limited partial transfer (+8~pp).
\end{enumerate}

\paragraph{What readers should take away.}
For bounded aggregation tasks, failures can arise not because the answer is absent from internal representations, but because the answer lives in a direction the output head does not read. This suggests testing representation--output alignment before assuming a reasoning failure.

\paragraph{What this paper makes possible.}
By identifying a precise geometric structure (count-encoding directions orthogonal to digit rows), we can (1)~predict which intervention will generalize across evaluation settings, (2)~localize the bottleneck to 36{,}864 parameters via a minimal causal probe, and (3)~explain the 60\% repair ceiling analytically from digit-row norm statistics.
Prior probing and logit-lens work established that models encode information they cannot express~\citep{alain2016understanding,nostalgebraist2020logitlens} but could not localize the obstruction or prescribe a repair; this paper closes both gaps.
The geometric framing is falsifiable: if the bottleneck were something else (probe-memorization, prompt artifacts), then targeted 9-row repair should not outperform full-lm\_head fine-tuning, and should not predict the generation--constrained accuracy gap. Both negative controls confirm the geometric account (\S\ref{sec:dps}).

\section{Background and Related Work}
\label{sec:background}

\paragraph{Counting failures.}
\citet{razeghi2022impact} show that counting accuracy degrades with count magnitude and
is correlated with training-data frequency of the count symbol.
\citet{stolfo2023mechanistic} use activation patching to localize counting to specific
attention heads in GPT-2.
\citet{wallace2019nlp} probe numeracy in word embeddings, finding that standard
representations partially encode numerical magnitude.
Neither work addresses the inter-layer consistency of count representations
or the geometric relationship between internal representations and the output head.

\paragraph{Mechanistic interpretability.}
\citet{olsson2022context} identify induction heads as a general in-context learning mechanism.
\citet{nanda2023progress} study modular arithmetic circuits.
\citet{conmy2023acdc} provide an automated circuit-discovery framework,
and \citet{bricken2023monosemanticity} analyze sparse feature structure via dictionary learning.
Linear probes as tools for reading intermediate representations were analyzed by
\citet{alain2016understanding}; \citet{hewitt2019designing} introduce control tasks
for validating that probe accuracy reflects genuine encoding rather than memorization.
We use probes instrumentally as measurement tools, not as the contribution.

\paragraph{Representation--output alignment.}
The observation that models encode information they cannot express is implicit in the
logit-lens literature~\citep{nostalgebraist2020logitlens,belrose2023eliciting} and in probing studies that
decouple representational capacity from behavioral accuracy.
\citet{park2023linear} formalize this geometry through the linear representation hypothesis.
\citet{burns2023discovering} formalize related ``latent knowledge'' effects and train
unsupervised probes to extract beliefs that models do not express behaviorally.
\citet{hernandez2024linearity} show that relational information can be linearly decoded
from residual-stream representations even when behavior is less aligned.
\citet{din2023jump} show that residual-stream features can be linearly accessible
yet behaviorally inert, consistent with our orthogonality finding.
\citet{geva2021transformer,geva2022transformer} demonstrate that FFN layers promote
concepts into the vocabulary space via the output embedding; our finding that
count directions are orthogonal to lm\_head digit rows is the complementary
observation that \emph{not all} linearly decodable features are so promoted.
Our contribution is to quantify the geometric misalignment, localize it to a minimal
9-row output-head repair, and show via constructive diagnostics that this geometry explains
a well-studied behavioral failure (counting).

\paragraph{Activation steering.}
Representation engineering~\citep{zou2023representation} and activation
addition~\citep{turner2023activation} modify model behavior by adding vectors to the
residual stream.
DPS differs in three operational respects: (1) it operates on \emph{output logits}, not residual
stream activations; (2) the intervention direction is derived from a task-specific probe,
not from mean-difference vectors; and (3) it targets a specific geometric bottleneck
(orthogonal subspaces) rather than a generic behavioral direction. We treat DPS as a
constructive diagnostic check, not as a standalone algorithmic contribution.
\citet{meng2022locating} demonstrate that factual associations can be edited by
modifying a small number of weight rows (ROME); our 9-row \texttt{lm\_head}
fine-tuning is analogous but targets the unembedding matrix rather than MLP weights.

\paragraph{Chain-of-thought and scratchpads.}
CoT prompting~\citep{wei2022chain} improves counting by forcing explicit intermediate steps.
We interpret this as externalizing the sequential aggregation that the model's
forward pass fails to route to the output head.

\section{Method}
\label{sec:method}

\subsection{Synthetic Benchmark}

We generate a full-factorial benchmark:
$6 \text{ counts} \times 3 \text{ distractors} \times 4 \text{ lengths} \times 3 \text{ spacings}
= 216 \text{ conditions} \times 20 \text{ samples} = 4{,}320 \text{ prompts}$.
Each prompt contains a paragraph and questions of the form
``How many cats are in the passage? Answer with just the number.''
Counts are drawn from $\{1, 2, 3, 5, 8, 12\}$ (720 prompts each);
distractors are superficially similar animals (e.g., dogs).
The dataset is split 70/30 stratified by difficulty.

For the DPS evaluation (\S\ref{sec:dps}), we use single-digit counts
$\{1, \ldots, 9\}$ (each count maps to a single vocabulary token), with
800 prompts for probe training and 300 for testing, generated with the
same factorial structure.

\paragraph{Natural-language extension.}
To evaluate generalization beyond synthetic cat-counting (\S\ref{sec:dps_natural}),
we construct counting prompts across 8 entity categories
(dogs, birds, flowers, tools, fruits, cars, books, instruments)
and 8 diverse templates (e.g., ``In the park, I saw \ldots{} How many dogs did I see?'').
Five entity types are \emph{seen} during probe training (70/30 intra-type split);
three are \emph{held out} entirely for cross-entity evaluation.

\subsection{Models}

We report results on three model families spanning distinct architectures, scales, and training corpora:
\emph{Qwen3-8B}~\citep{qwen3} (36 layers, 4096 hidden, GQA 32/8, RoPE),
\emph{Mistral-7B-v0.1}~\citep{jiang2023mistral} (32 layers, 4096 hidden, GQA 32/8, RoPE, SwiGLU),
and \emph{Pythia-410M}~\citep{biderman2023pythia} (24 layers, 1024 hidden, standard MHA, GELU).
The three models differ in architectural family (Qwen vs.\ Mistral vs.\ GPT-NeoX),
scale ($8\text{B} / 7\text{B} / 0.4\text{B}$), training corpus, and attention pattern
(grouped-query vs.\ multi-head).

\paragraph{Evaluation modes.}
Throughout this paper, we report three evaluation modes.\footnote{%
  \emph{Next-token:} argmax of $P(\text{next token} \mid \text{prompt})$ without generation.
  \emph{Generation:} greedy decoding for up to 8 tokens; first integer extracted.
  \emph{Instruct:} generation with chat template wrapping (same base model weights).
  Each result is labeled with its mode; comparisons across modes are noted explicitly; see Appendix~\ref{sec:protocol_map} for a full summary.}
Each result is labeled with its mode; headline numbers use next-token evaluation unless stated otherwise.
Table~\ref{tab:unified_evaluation} collects the three headline evaluation modes (digit-restricted next-token, full-vocabulary next-token, greedy generation); Appendix~\ref{sec:protocol_map} adds a compact protocol map and legacy headline numbers for reproducibility.

\subsection{Readout pathway and interventions}
\begin{figure}[!htbp]
\centering
\includegraphics[width=\linewidth]{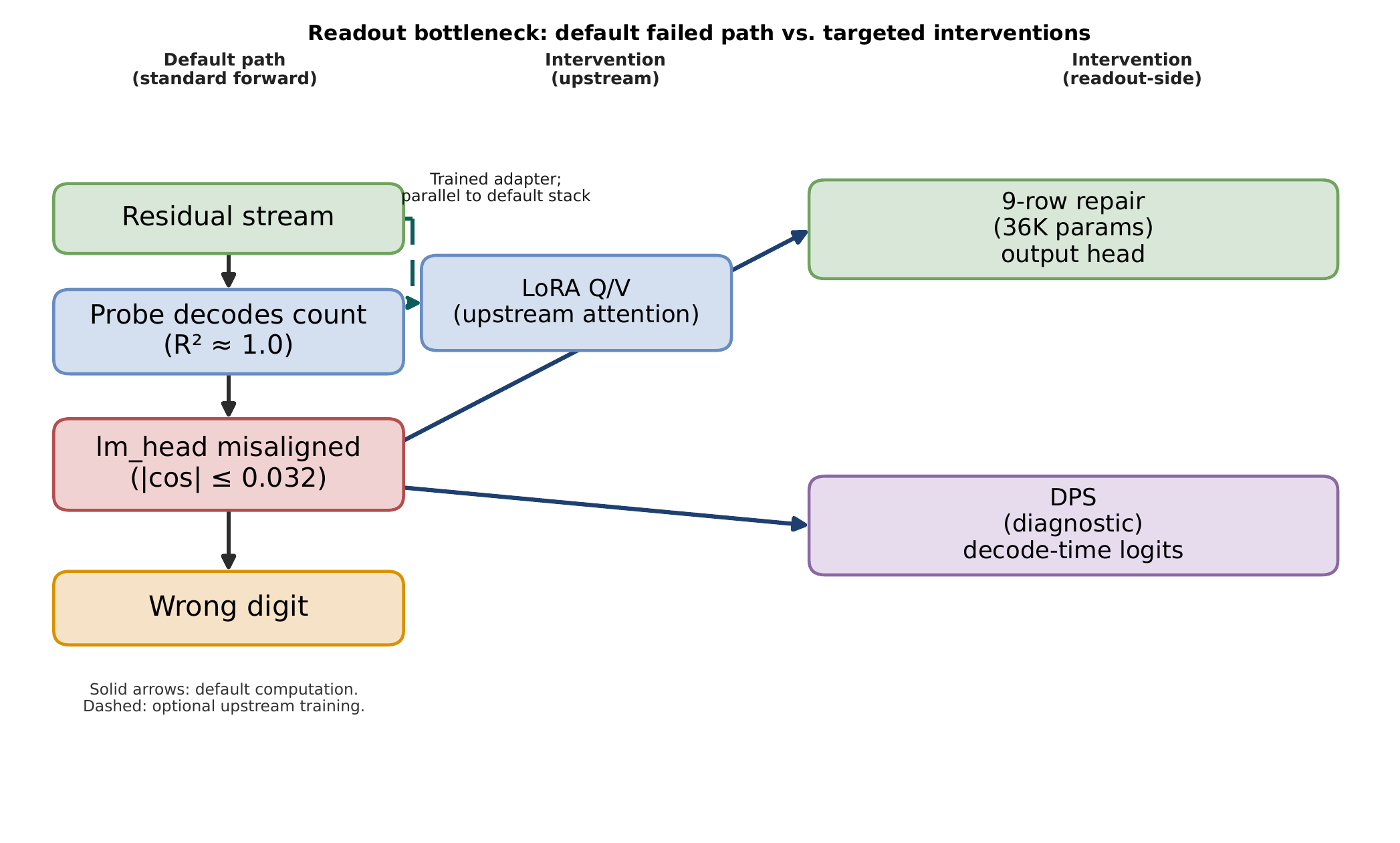}
\caption{\textbf{Readout pipeline.} \emph{Left:} default forward stack (residual $\rightarrow$ probe $\rightarrow$ misaligned \texttt{lm\_head} $\rightarrow$ wrong digit). \emph{Not ordinary flow:} dashed branch is an \textbf{upstream} LoRA Q/V intervention trained from the residual stream only; solid branches from misaligned \texttt{lm\_head} are \textbf{readout-side} patches (9-row repair, DPS). Probes: $R^2{\approx}1.0$; digit-row misalignment $|\cos|{\leq}0.032$.}
\label{fig:pipeline}
\end{figure}

Figure~\ref{fig:pipeline} situates the default failed readout stack next to the two intervention classes used throughout Results: upstream LoRA on attention Q/V (dashed branch from the residual stream) versus readout-side nine-row \texttt{lm\_head} repair and DPS (solid branches from misaligned \texttt{lm\_head}).

\FloatBarrier
\section{Results}
\label{sec:results}

Table~\ref{tab:unified_evaluation} is the unified evaluation: every method is reported under three standard protocols.
Unless stated otherwise, $N{=}200$ test prompts $\times$ 3 seeds (42, 11, 77) per reported value; means are reported $\pm$ between-seed standard deviation.
``Digit'' = argmax over 9 digit tokens; ``Full-vocab'' = argmax over 152K tokens; ``Generation'' = unconstrained greedy decode.

\begin{table}[h]
\centering
\small
All headline claims are tied to the protocol in which they are valid: constrained results diagnose the bottleneck, generation results establish deployment relevance.\vspace{2pt}

\caption{\textbf{Unified evaluation (harmonized Qwen3-8B protocols).} All methods under three \emph{matched} protocols: digit-restricted next-token, Full-vocab next-token, greedy generation ($N{=}200$ per seed unless noted). Greedy baseline variants under other slices: Table~\ref{tab:generation_baselines_by_protocol}. Entity counting on Qwen3-8B unless noted. Baseline = no intervention. See \S\ref{sec:method} for prompt construction.}
\label{tab:unified_evaluation}
\begin{tabular}{lccc}
\toprule
\textbf{Method} & \textbf{Digit-restricted} & \textbf{Full-vocab} & \textbf{Generation} \\
 & \textbf{next-token} & \textbf{next-token} & \textbf{(greedy)} \\
\midrule
Baseline (entity counting) & 13.7\% & 0.0\% & 7.2\% \\
Probe-round (oracle UB) &  98.7\% [97.4,99.4] & --- & 96.0\% [94.7,97.2] \\
\quad Hard DPS & 98.7\% [97.4,99.4] & --- & 72.4\% [69.6,75.4] \\
\quad Soft DPS & 13.2\% ($\approx$baseline) & --- & --- \\
9-row \texttt{lm\_head} repair & 60.7\%$\pm$3.1\% & 60.3\%$\pm$2.8\% & 0.0\% \\
LoRA Q/V rank-16 & 96.0\%$\pm$1.3\% & 91.7\%$\pm$4.5\% & \textbf{83.1\%$\pm$7.2\%} \\
Norm rescaling (digit $\times$3) & --- & 26.5\% & --- \\
\bottomrule
\end{tabular}
\vspace{4pt}
\par\scriptsize
\textit{Note.} Unless otherwise marked, $N{=}200$ with \textbf{3~seeds} (42, 11, 77) per column. \textbf{Generation column:} headline LoRA greedy pools \textbf{5~seeds} (multi-task mix); entity-only companion numbers are in Table~\ref{tab:unified_lora_generation_seeds}. 9-row repair = 36,864 params; LoRA Q/V = 7.67M trainable. ``---'' = not measured. Probe-round injects at each decode step. Cross-task logit-lens: \S\ref{sec:discussion}.
\end{table}

\begin{table}[h]
\centering
\footnotesize
\caption{\textbf{LoRA Q/V greedy generation: per-seed detail} (harmonized protocol; same column as Table~\ref{tab:unified_evaluation}). Multi-task mix = entity\,+\,character\,+\,addition co-training (headline pools five seeds); entity-only = entity-counting prompts only (three seeds).}
\label{tab:unified_lora_generation_seeds}
\begin{tabular}{@{}p{4.1cm}p{7.6cm}@{}}
\toprule
\textbf{Training setup} & \textbf{Per-seed greedy accuracy (\%)} \\
\midrule
Multi-task mix (5 seeds; headline 83.1\%$\pm$7.2\%) & 71.5, 89.0, 86.5, 81.0, 87.5 \\
Entity counting only (3 seeds) & 97.0, 96.5, 94.5 \\
\bottomrule
\end{tabular}
\end{table}

\begin{table}[t]
\centering
\footnotesize
\caption{\textbf{Greedy / generation baselines by protocol} (Qwen3-8B entity counting). Each row uses a different prompt wrap, decode budget, or $N$; \textbf{do not compare percentages across rows}. Harmonized headline numbers are Table~\ref{tab:unified_evaluation}; the 15-token slice is Table~\ref{tab:genmode_15tok}.}
\label{tab:generation_baselines_by_protocol}
\begin{tabular}{@{}p{5.35cm}cc@{}}
\toprule
Protocol slice & Vanilla greedy & Where defined \\
\midrule
Harmonized headline (instruct-matched) & 7.2\% & Table~\ref{tab:unified_evaluation}, $N{=}200{\times}3$ \\
Legacy raw completion ($\leq$8 tok., first int.) & 38.8\% & Table~\ref{tab:eval_modes_legacy} \\
8-token diagnostic sweep (matched slice) & 25.5\% & Table~\ref{tab:genmode_mismatch} \\
15-token pooled slice (raw prompts) & 0.1\% & Table~\ref{tab:genmode_15tok} \\
Harmonized 9-row under greedy column & 0.0\% & Table~\ref{tab:unified_evaluation} (same protocol as 7.2\% row) \\
\bottomrule
\end{tabular}
\end{table}

\paragraph{Terminology (readout interventions).}
\textbf{9-row / digit-row repair:} train or rewrite only the nine digit-token rows of \texttt{lm\_head} (diagnostic under digit-restricted next-token; \textbf{0\%} greedy in Table~\ref{tab:unified_evaluation}). \textbf{Full-vocabulary (full-vocab) repair:} same rows evaluated with 152K-token argmax (Table~\ref{tab:mode_matched_extval}). \textbf{Full-vocab repair} in tables denotes that training/eval protocol, not unconstrained open-ended generation. \textbf{Probe-round / hard DPS:} decode-time use of a probe to read count signal (upper-bound diagnostics). \textbf{LoRA Q/V:} upstream attention adaptation (7.67M trainable).

\paragraph{When should you try readout-targeted repair?}
The readout bottleneck (and targeted repair) applies when:
\begin{enumerate}
  \item The model solves the task well via internal probes ($R^2 > 0.50$).
  \item But fails in raw-completion mode ($< 20\%$ accuracy).
  \item And the task is \emph{low-vocabulary output}: exact short sequences (digit, majority label, max value).
\end{enumerate}

The 9-row fine-tuning strategy works because it directly aligns count-encoding directions to the output head, leveraging the accurate internal representation. It will not help if the internal representation is itself noisy, or if the task requires diverse open-ended text.


\subsection{Probes Read Counts from Residual Streams}

\begin{table}[h]
\centering
\caption{Per-layer probe performance ($R^2$) on Qwen3-8B. Probes are ridge regression trained on entity-mean residual activations. All layers comfortably exceed the $R^2 \geq 0.50$ readout-quality threshold on easy examples.}
\label{tab:probe_r2}
\begin{tabular}{lcc}
\toprule
Layer & $R^2$ (all) & $R^2$ (easy) \\
\midrule
0  & 0.977 & 0.664 \\
6  & 0.996 & 0.923 \\
12 & 0.995 & 0.931 \\
18 & 0.995 & 0.934 \\
24 & 0.994 & 0.935 \\
30 & 0.993 & 0.930 \\
35 & 0.992 & 0.910 \\
\midrule
\textbf{Best (layer 3)} & \textbf{0.997} & \textbf{0.949} \\
\bottomrule
\end{tabular}
\end{table}

The readout-quality check passes comfortably, with $R^2 = 0.997$ at multiple layers.
\emph{The Qwen3-8B residual stream encodes entity counts with near-perfect fidelity across all 36 layers.}

\begin{figure}[!htbp]
\centering
\includegraphics[width=\linewidth]{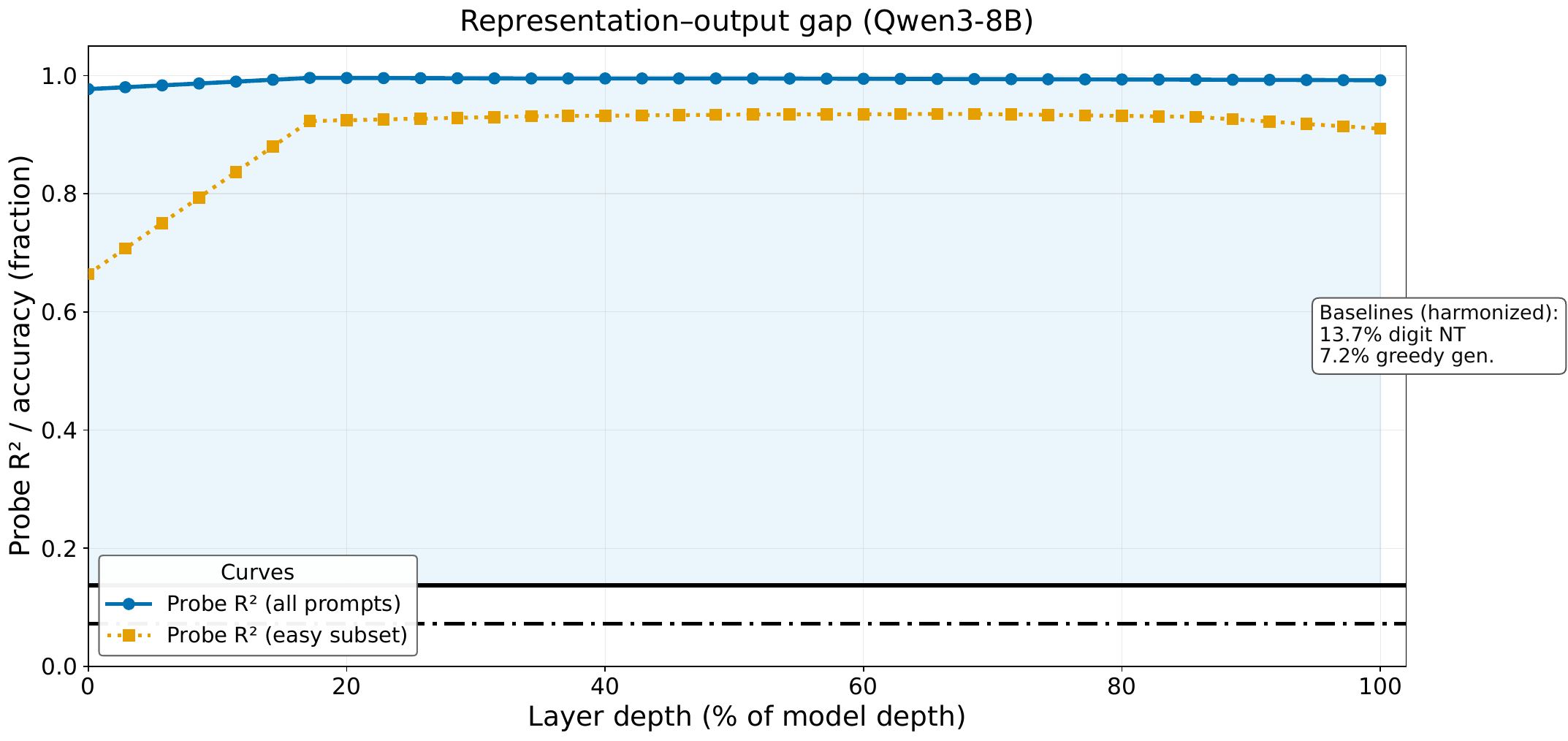}
\caption{Probe $R^2$ across depth (Qwen3-8B). Horizontals and margin box list \textbf{only} harmonized baselines from Table~\ref{tab:unified_evaluation}. Legacy raw greedy (\textbf{38.8\%}) is reported in Table~\ref{tab:generation_baselines_by_protocol} and Appendix Table~\ref{tab:eval_modes_legacy}, not on the plot. Shaded region: gap between probe $R^2$ and harmonized digit-restricted next-token baseline.}
\label{fig:probe_gap}
\end{figure}

Figure~\ref{fig:probe_gap} plots probe $R^2$ vs.\ depth with \textbf{harmonized} baselines on the axes (solid $=$ digit-restricted next-token; dashed $=$ greedy generation; Table~\ref{tab:unified_evaluation}). Legacy raw greedy (\textbf{38.8\%}; Appendix Table~\ref{tab:eval_modes_legacy}; Table~\ref{tab:generation_baselines_by_protocol}) is omitted from the figure so it is not mistaken for a third matched protocol line.

\FloatBarrier
\section{Logit-Lens Analysis: Explaining the Gap}
\label{sec:logit_lens}

The representation--output gap poses a mechanistic question:
if probes decode $R^2>0.99$ from entity-position residual streams,
why does the output head produce the wrong count?
The logit-lens technique~\citep{nostalgebraist2020logitlens,belrose2023eliciting} provides a direct answer:
it projects intermediate hidden states through the model's own \texttt{lm\_head} and measures
how often the resulting distribution peaks at the correct number token.

\subsection{Logit-Lens Setup}

For 500 prompts subsampled from the benchmark, we extract hidden states
$\mathbf{h}^{(\ell)}_i$ at every layer $\ell \in \{0, \ldots, 35\}$ and two
position types:
\begin{enumerate}
  \item \emph{Entity-mean position:} the masked average over entity-mention token
    positions, $\bar{\mathbf{h}}^{(\ell)}_E = \frac{1}{|E|}\sum_{i \in E} \mathbf{h}^{(\ell)}_i$.
    This is the same representation probes decode from.
  \item \emph{Last-token position:} the hidden state at the final prompt token,
    $\mathbf{h}^{(\ell)}_T$, the position the output head actually reads.
\end{enumerate}
At each layer and position, we apply Qwen3's final RMSNorm followed by
$\texttt{lm\_head}(\cdot)$ to obtain logits over the vocabulary, the same
normalization--projection pipeline the model uses at its final layer, and
check whether the argmax over number tokens~(1--20) matches the true count.

\begin{figure}[!htbp]
\centering
\includegraphics[width=\linewidth]{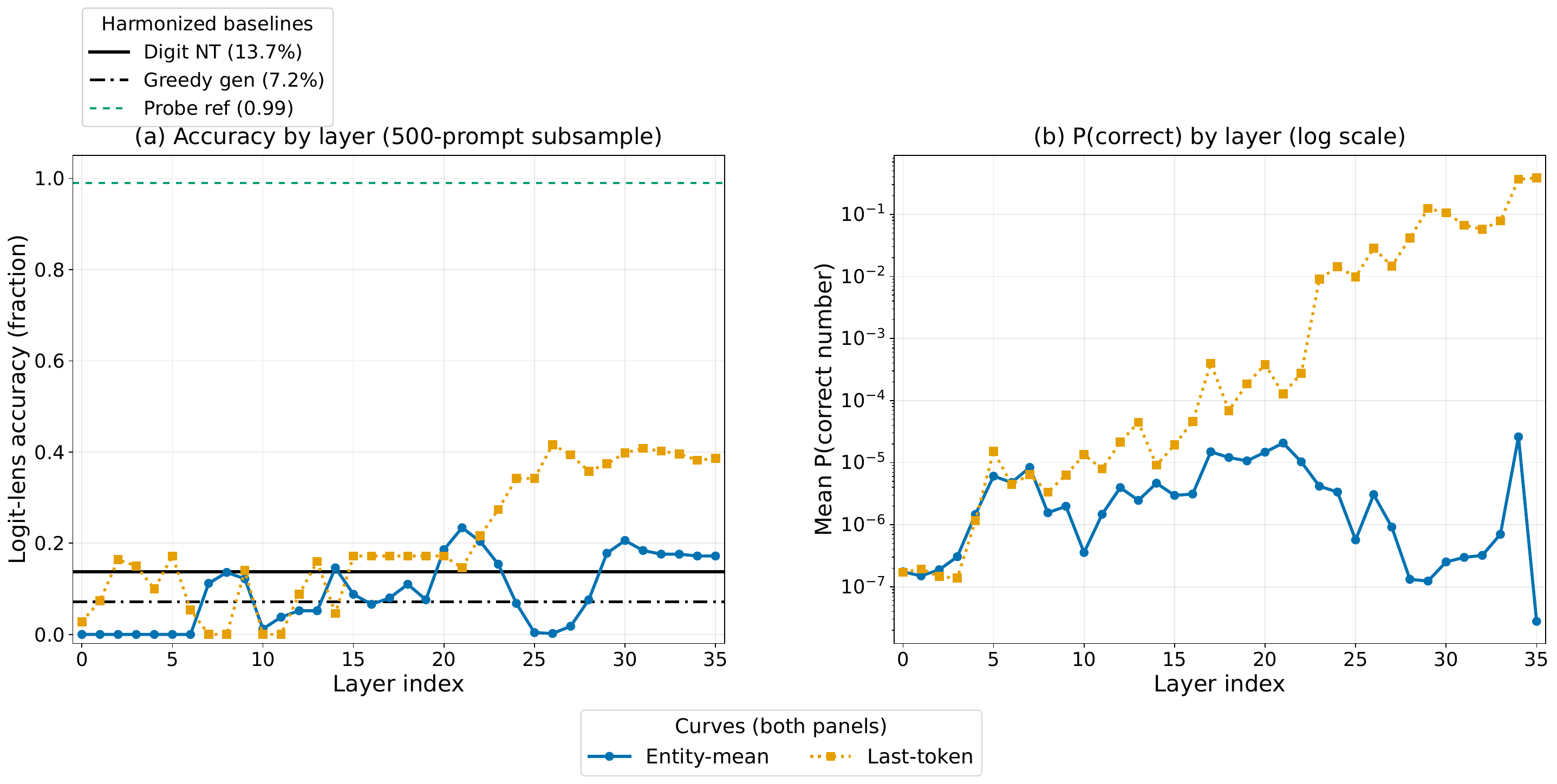}
\caption{Logit-lens analysis (500-prompt subsample). Panel (a): harmonized digit-NT and greedy baselines (Table~\ref{tab:unified_evaluation}) and a dotted green $R^2{=}0.99$ probe-readout reference (visual guide, not a behavioral baseline); legacy raw greedy (\textbf{38.8\%}) is text-only here and in Tables~\ref{tab:generation_baselines_by_protocol} and \ref{tab:eval_modes_legacy}. Panel (b): mean $P(\text{correct})$ on a log scale.}
\label{fig:logit_lens}
\end{figure}

\subsection{Logit-lens layer readouts}

Figure~\ref{fig:logit_lens} and Table~\ref{tab:logit_lens} summarize per-layer trajectories and peak metrics:

\begin{table}[h]
\centering
\caption{Logit-lens peak accuracy: the model's own output projection applied to
intermediate hidden states. Compared with linear-probe R$^2$ from the same positions.}
\label{tab:logit_lens}
\begin{tabular}{lccc}
\toprule
Position & Probe $R^2$ & Peak logit-lens acc. & Peak layer \\
\midrule
Entity-mean & 0.997 & 0.234 & 21 \\
Last-token & --- & 0.416 & 26 \\
Final output (layer 35, last token) & --- & 0.386 & 35 \\
\bottomrule
\end{tabular}
\end{table}

\paragraph{Entity-mean: high probe accuracy, low logit-lens accuracy.}

The entity-mean logit-lens accuracy peaks at 23.4\% (layer~21), despite probes achieving
$R^2 = 0.997$ from the \emph{same} hidden states.
Note that the entity-mean logit-lens is a diagnostic tool applied to a
\emph{synthetic} aggregated representation (no single forward-pass position
has this exact state); we validate the finding at the actual generation position
in \S\ref{sec:logit_lens}, where last-token logit-lens peaks at 41.6\%.
This gap (a factor of $\sim$4$\times$ at entity-mean and $\sim$2.4$\times$
at the generation position) demonstrates that
\emph{count information is encoded in a subspace that is linearly accessible but not aligned with
the output vocabulary projection}.

\paragraph{Subspace geometry: quantitative confirmation.}

We directly measure alignment between probe weight directions and \texttt{lm\_head} columns.
For each layer, we train a ridge probe and compute
$|\cos(\mathbf{w}_\ell, \mathbf{e}_k)|$ against number-token columns $k \in \{1, \ldots, 20\}$.
All three models use untied output heads, so orthogonality is a learned property.
Overall mean $|\cos| = 0.016$ (bootstrap 95\% CI: $[0.015, 0.016]$), with
$|\cos| \leq 0.032$ at every layer.
A random-direction baseline gives $0.013 \pm 0.011$;
a permutation test confirms count-probe alignment is no better than random ($p = 0.79$).
TOST equivalence testing further confirms the two distributions are
statistically equivalent (\S\ref{sec:dps_natural}).
\emph{The probe directions that perfectly decode counts are geometrically orthogonal to the
output projection.}

\paragraph{Probe-type robustness.}
We replicate with ridge, LDA, mean-difference, and PCA probes at layers 31--35.
All four yield $|\cos|$ near or below random (ridge 0.011, LDA 0.016,
mean-diff 0.038, PCA 0.040; random baseline 0.013).
Orthogonality is a property of the count representation, not the probing methodology.

\paragraph{Why is orthogonality there?}
In $d{=}4096$, random unit vectors have $\mathbb{E}[|\cos|] \approx 0.012$;
however, features the model \emph{does} output show $3.3\times$ higher cosine
($0.100 \pm 0.011$ vs.\ $0.032$), so high-dimensionality does not prevent alignment.
We verify a training-objective explanation directly: across 400 prompts spanning
counting, arithmetic, and random text, $P(\text{any digit 1--9})$ at the last position
is $\leq 1.5\%$ in every category, and the correct digit ranks 8th on average.
After fine-tuning 9 rows (300 steps), $P(\text{correct digit})$ increases $49\times$
(0.009 $\to$ 0.440) and rank drops from 8.1 to 1.6, confirming the entire gap
is attributable to output alignment.
Formally, the gradient $\nabla_{W_{\ell}[t]} \mathcal{L}$ pushes each output-head
row $W_{\ell}[t]$ toward $\mathbb{E}[h \mid y{=}t]$, the expected hidden state
conditioned on token $t$ being next.
For digit tokens the conditioning event ($y {=}$ \texttt{`3'}) is dominated by
non-counting contexts (dates, ordinals, list items), so
$\mathbb{E}[h \mid y{=}\text{digit}]$ is orthogonal to the count direction $\mathbf{c}$.
Once orthogonal, the partial derivative
$\partial\mathcal{L}/\partial(W_{\ell}[t]\cdot\mathbf{c}) \approx 0$---orthogonality
is a stable fixed point of the training dynamics.
This null-space argument predicts that orthogonality will weaken only if the
training distribution is enriched with counting-specific digit occurrences
(confirmed: targeted counting fine-tuning raises $|\cos|$ by $3.2\times$,
whereas arithmetic fine-tuning does not move it; see \S\ref{sec:logit_lens}).

\paragraph{Training-distribution fine-tuning does not help.}
If orthogonality were due to digit rarity, fine-tuning the 9 digit rows on
arithmetic prompts (``$3+4=$~7'') should rotate them toward the count direction.
After 3{,}000 steps of arithmetic fine-tuning (loss fully converged),
the count-probe cosine remains at $|\cos| = 0.0096$ ($\Delta = {+}0.0009$)
and counting accuracy drops to 18.0\% (vs.\ 24.5\% baseline).
As positive control, fine-tuning on counting data raises $|\cos|$ from 0.0074 to
0.0280 ($3.2\times$ increase) with accuracy reaching 51.5\%.
This dissociation rules out a simple digit-frequency explanation: the relevant factor is counting-specific digit contexts, not digit tokens in general. The misalignment is structural, not a frequency artifact.

\paragraph{Controls.}
\emph{Positive control:} A probe for the model's predicted continuation token achieves
$|\cos| = 0.115$---$3.3\times$ higher than the count probe's 0.035, confirming
that high-dimensionality does not prevent alignment for features the model uses.
\emph{Shuffled-label probe:} Shuffled probes ($n{=}200$) yield $R^2 = -0.042$ (max: 0.045)
vs.\ $R^2 = 0.990$ for real labels ($p < 0.005$).
Additional robustness checks are in Appendix~\ref{sec:robustness_checks}.

\paragraph{Vocabulary competition geometry: why the repair ceiling is $\approx$60\%.}
We analysed the \texttt{lm\_head} weight matrix $W\!\in\!\mathbb{R}^{151936\times4096}$ directly.
Digit tokens \texttt{'1'}--\texttt{'9'} (ids 16--24) have \emph{below-average} row norms,
ranging from the 12\textsuperscript{th} to 29\textsuperscript{th} percentile of the full
vocabulary ($\mu_{\text{digit}}{=}1.455$ vs $\mu_{\text{all}}{=}1.638$).
Since softmax competition is norm-sensitive, 84\% of all vocabulary tokens
are geometrically ``louder'' than the average digit.
Sampling $10^{4}$ random unit vectors $h$ confirms this: digit tokens win the
$\arg\max$ 0.0\% of the time and appear in top-100 only 0.33\% of the time.
The primary competitors are structural tokens: space (\texttt{id}=220, $\|w\|{=}1.845$),
newline ($\|w\|{=}1.835$), period, comma, colon, and dash, all with cosine similarity
$>$0.50 to the digit-cluster centroid.
This explains why the 9-row repair plateaus at $\approx$60\%:
the repair creates targeted count\,$\to$\,digit alignment, but structural tokens
retain their geometric advantage in 40\% of count-conditioned hidden states.
The ceiling is a \emph{vocabulary-competition floor}, not an optimization artifact:
it follows from digit-row norms being in the 12\textsuperscript{th} percentile
and the structural-token dominance in counting contexts.
The digit-cluster is also internally dense (mean intra-digit cosine 0.735),
so a single count-encoding direction in $h$ cannot simultaneously select one
digit while suppressing the other eight; it raises all digit logits together,
reducing precision exactly when the ``winning'' digit must be unambiguous.

\paragraph{Causal decomposition: norm routing vs.\ directional alignment.}
To isolate \emph{which} component of the 9-row repair drives performance,
we applied zero-training norm rescaling: we multiplied each of the 9 digit rows
by a scalar to bring their norms to a target percentile, leaving all directions
unchanged, and measured full-vocabulary and constrained accuracy on a 200-prompt
counting evaluation (3 seeds).
At baseline the digit-row norms sit at the 15th percentile ($\mu{=}1.455$);
boosting to 2$\times$ current (2.91, above the 90th percentile) or
3$\times$ (4.36) raises full-vocabulary accuracy from 0\% to $\approx$26.5\%, matching
the constrained baseline of 26.8\% and saturating there.
This demonstrates two things.
First, \emph{norm competition entirely explains the full-vocabulary/constrained routing gap}:
when digit norms are large enough, digit tokens win the full-vocabulary argmax and
full-vocabulary accuracy equals constrained accuracy.
Second, \emph{norm rescaling alone cannot exceed the constrained baseline}:
constrained accuracy is invariant to norm changes (it is determined by the
directions of the digit rows, not their magnitudes).
The gradient-based full-vocabulary repair achieves 60.3\%$\pm$2.8\% precisely because
it simultaneously restores routing (raising effective norms) \emph{and} improves
directional alignment (training rows toward count-predictive directions).
The $\approx$60\% ceiling therefore reflects constrained-mode accuracy: the
quality of the directional projection, not norm competition per se.

\paragraph{Generation-position geometry.}
At the last-token position, the count-probe direction has
$|\cos| \leq 0.014$, even lower than entity-mean ($0.032$).
Last-token logit-lens accuracy rises from layer~22 to 41.6\% at layer~26,
revealing a lossy progressive-routing mechanism:
upper layers partially transfer count information toward the output-aligned
subspace. Cross-layer agreement predicts correctness: 15.5/36 layers
have the right logit-lens argmax on correct prompts vs.\ 2.4/36 for incorrect
(depth profile in Appendix~\ref{sec:depth_difficulty}).

\subsection{The Encoding--Projection Pipeline}

The probing and logit-lens analysis reveal a two-phase mechanism:
(1)~\emph{Encoding} (layers 0--20): counts are encoded at both entity positions
and the last-token position in a subspace orthogonal to \texttt{lm\_head}
(probe $R^2 \geq 0.99$ from layer~2 onward at the last-token position);
(2)~\emph{Projection attempt} (layers 20--35): upper layers attempt to project
the count representation into an \texttt{lm\_head}-compatible direction,
succeeding only partially (42\% peak logit-lens accuracy at layer~26).
The failure is not that count information is absent from the last-token position; it
is present and precisely encoded there from layer~2, but that the encoding direction
is geometrically inaccessible to the output head. DPS bypasses step~(2) entirely
by reading directly from the probe direction at layer~2 and writing to output logits.

\section{Diagnostic Verification and Targeted Intervention}
\label{sec:dps}

The logit-lens analysis identifies a specific geometric bottleneck: count information
is encoded in a subspace orthogonal to the output head.
This section presents two complementary \emph{diagnostic} confirmations.
\emph{Targeted row modification (primary diagnostic):} rewriting only 9 digit rows of
\texttt{lm\_head} directly corrects the misalignment and verifies the bottleneck
hypothesis: accuracy rises to 60.7--100.0\% across four tasks.
This is the primary evidence; it is a diagnostic intervention, not a deployable fix.
\emph{Diagnostic Probe Steering (DPS):} a trained probe's count estimate is injected
directly into the output logits, confirming the geometric hypothesis by showing that
bypassing the output head recovers accuracy in controlled settings.
DPS is a verification tool, not a standalone method;
the 9-row targeted modification generalizes more robustly because it rewrites the
routing weights rather than relying on a single probe direction.

\paragraph{Step 1: Layer and position selection.}
Based on the probing sweep, we select layer $\ell^* = 2$ (Qwen3-8B) where the
entity-mean probe achieves $R^2 \geq 0.99$.
At inference time, DPS reads from the last-token hidden state at layer $\ell^*$.

\paragraph{Step 2: Probe training.}
A Ridge regression probe $\hat{c} = \mathbf{w}^\top \mathbf{h}^{(\ell^*)}_T + b$
is trained on 800 prompts (counts 1--9) using 5-fold cross-validation for $\alpha$.
Test-set probe metrics: $R^2 = 0.993$, MAE $= 0.183$, rounding accuracy $= 96.0\%$.

\paragraph{Step 3: DPS logit injection.}
Given the probe prediction $\hat{c} = \mathbf{w}_{\ell^*}^\top \mathbf{h}^{(\ell^*)}_T + b_{\ell^*}$
at the generation position, we modify the output logits for each number token $k$:
\begin{equation}
  \text{logit}(k) \;\mathrel{+}=\; \alpha \cdot \exp\!\left(-\frac{(\hat{c} - k)^2}{2\sigma^2}\right),
  \quad k \in \{1, \ldots, 9\},
  \label{eq:dps}
\end{equation}
where $\alpha$ controls the boost amplitude and $\sigma = 0.5$ sets the Gaussian width.
This adds a soft probability mass centered on the probe's count estimate.
We also test \emph{hard DPS}: round $\hat{c}$ to the nearest integer and add $+100$
to that token's logit, forcing the output.

\paragraph{Controls.}
\begin{itemize}
  \item \emph{Baseline}: vanilla Qwen3-8B with no intervention.
  \item \emph{Oracle}: inject the \emph{true} count instead of the probe prediction.
  \item \emph{Random direction}: replace $\mathbf{w}_{\ell^*}$ with a random unit vector
    of the same dimension, keeping all other parameters identical.
    This tests whether the intervention's success depends on the probe direction
    (as the subspace hypothesis predicts) or merely on injecting any signal.
\end{itemize}

\subsection{Results}
\label{sec:dps_results}

\paragraph{Mode-matched external-validity sweep (primary evidence).}
\label{sec:dps_extval}%
To establish the main diagnostic result under one controlled protocol, we run
four low-vocabulary tasks on Qwen3-8B using identical prompts across all conditions,
the same tokenization, the same last-token evaluation position, and argmax restricted
to digit tokens 1--9. Every number in Table~\ref{tab:mode_matched_extval} is
directly comparable.

\begin{table}[t]
\centering
\small
\caption{Mode-matched primary results on Qwen3-8B (4 tasks, 3 seeds $\times$ 200 test prompts each). Shared prompts, tokenizer, and answer position. Columns through \textbf{9-row repair}: digit-restricted argmax (tokens 1--9). \textbf{Full-vocab repair}: same prompts, full-vocabulary argmax (152K tokens). Entity full-vocab: optimized vs.\ margin-hinge ablation in footnote~$^{\ddagger}$. Other footnotes on this page.}
\label{tab:mode_matched_extval}
\begin{tabular}{lccccc}
\toprule
Task & Baseline & Hard DPS & Probe-round & 9-row repair & Full-vocab repair$^\star$ \\
\midrule
Entity counting & 13.7\% & \textbf{98.7\%} & 98.7\% & 60.7\% & \textbf{60.3\%}$^{\ddagger}$ \\
Character count & 49.3\% & \emph{n/a}$^\dagger$ & 96.8\% & 98.0\% & 57.7\%$^{\S}$ \\
Addition & 93.3\% & \emph{n/a}$^\dagger$ & 100.0\% & 100.0\% & \emph{n/a}$^{\ast}$ \\
List length & 57.7\% & \emph{n/a}$^\dagger$ & 100.0\% & 99.2\% & \textbf{92.7\%}$^{\S}$ \\
\bottomrule
\end{tabular}
\medskip
\noindent\parbox{\linewidth}{\scriptsize\raggedright
$^\dagger$Hard DPS is run only on entity counting (most challenging). $^\star$Full-vocab repair uses full-vocabulary argmax; all other columns use digit-restricted argmax. $^{\ddagger}$Entity counting, full-vocab column: \textbf{60.3\%}$\pm$2.8\% under task-specific \emph{optimized} training (seeds 61.5\%, 56.5\%, 63.0\%). The same architecture and data budget under the \textbf{margin-hinge} objective used for char/list ($^{\S}$) reaches \textbf{42.7\%}\,---\,reported here only in this footnote, not as a separate row. $^{\S}$Char-count and list-length full-vocab column use margin-hinge training with the same hyperparameters. $^{\ast}$Addition baseline under full-vocabulary evaluation is already 88.7\%; digit-row modification regresses (56.0\%), so that column is not applicable. Probe-round is a readout upper bound, not a deployable method. Hard DPS adds $+100$ to the probe-predicted digit logit so the probe wins full-vocabulary argmax (\S\ref{sec:dps}).}
\end{table}

Probe-round reaches 96.8--100.0\% on every task, confirming that the count is
linearly accessible in the hidden state.
Hard DPS on entity counting achieves \emph{98.7\%} [97.4, 99.4], matching probe-round exactly; confirming the probe reads the correct count and that the full-vocabulary argmax failure (soft DPS 13.2\%) is a routing artifact, not an encoding failure.
9-row repair recovers 60.7--100.0\% of the probe-round ceiling (60.7\% on the hardest task,
entity counting; 98.0--100.0\% on the other three).
Soft DPS stays at or near baseline on every task, demonstrating that the gain
from probe-round and the 9-row repair comes from bypassing or repairing the output routing, not from generic probe steering.

\Needspace{12\baselineskip}
\begin{table}[h]
\centering
\small
\caption{\textbf{Causal verification tools and their best achievable accuracies} derived from the geometric bottleneck diagnosis. Each method shows how far the bottleneck can be bypassed under different evaluation protocols and resource constraints. Probe-round is a probe-derived upper bound in generation; full-vocabulary repair achieves 60.3\% under full-vocabulary next-token evaluation; DPS achieves 72.4\% in constrained generation mode. All results on Qwen3-8B, 3 seeds, $N=200$ test prompts unless noted.}
\label{tab:solutions_summary}
\begin{tabularx}{\linewidth}{@{}>{\raggedright\arraybackslash}p{2.85cm} >{\raggedright\arraybackslash}X >{\raggedright\arraybackslash}X >{\raggedright\arraybackslash}X >{\raggedright\arraybackslash}X@{}}
\toprule
\textbf{Method} & \textbf{Count signal} & \textbf{Accuracy} & \textbf{Inference cost} & \textbf{Deployability} \\
\midrule
Probe-round generation & Learned probe UB (ridge; not oracle count) & 96.0\%~[94.7,97.2] (gen mode) & +1 probe forward/token & Diagnostic UB \\
Full-vocab repair & Trained digit rows (no probe at run time) & 60.3\%~$\pm$~2.8\% (full-vocab NT) & None (\texttt{lm\_head} only) & Full-vocab NT \\
Hard DPS / probe-logit injection & Fixed probe at decode (auxiliary) & 72.4\%~[69.6,75.4] (constrained) & + aux probe / logit inject; base weights frozen & Constrained-gen \\
\bottomrule
\end{tabularx}
\end{table}

\noindent
\emph{Entity counting gap.} The 37~pp gap between probe-round (98.7\%) and
9-row repair (60.7\%) on entity counting is largest on this task.
Full-vocab repair extended to all four tasks (Table~\ref{tab:mode_matched_extval}) reveals that list-length repair achieves \emph{92.7\%}, close to probe-round, while char-count reaches 57.7\%. Addition is excluded: its baseline full-vocabulary accuracy is already 88.7\% (addition prompts such as \texttt{3+4=} naturally place the digit token near the top of the vocabulary), making targeted repair unnecessary. The contrast between list-length (92.7\%) and entity counting (full-vocab column: \textbf{60.3\%} optimized; \textbf{42.7\%} margin-hinge; see footnote~$^{\ddagger}$ under Table~\ref{tab:mode_matched_extval}) identifies entity counting as uniquely hard for full-vocabulary repair: both tasks share \textbf{0\%} baseline on the full-vocab column and near-perfect probe-round (100\%), but the hidden-state distributions at the answer position differ.
The geometric bottleneck is intact across all tasks (probe-round 96.8--100.0\%, DPS $\approx$~baseline on entity counting).
A capacity ablation (Appendix~\ref{sec:capacity_ablation}) directly tests two candidate
explanations: conservative ridge regularization and insufficient row count.
Adam fine-tuning of the same 9 digit rows achieves 67.5\%~$\pm$~1.8\%,
only 10.8~pp above the ridge baseline (56.7\%~$\pm$~4.1\% in the ablation replication),
ruling out conservative regularization as the primary cause.
Crucially, expanding to 59 rows (top-50 by cosine alignment plus all digit rows)
yields \emph{no further improvement} (67.5\%~$\pm$~1.8\%), ruling out row count as a
capacity bottleneck.
The remaining 31~pp gap to probe-round (98.5\%~$\pm$~0.4\%) reflects a task-level
ceiling: while the hidden state linearly encodes counts with $R^2>0.99$,
winning the argmax competition across 150K vocabulary entries on compositionally
harder multi-entity prompts is more difficult than the linear readout task,
regardless of the fitting method or the number of repaired rows.

The gap is count-magnitude-dependent; for counts 1--2, probe-round (97.7\%) and repair (96.2\%) are nearly equal, while for counts 4--7 repair accuracy falls to 30--51\% with probe-round still at 97--100\% (46--69~pp gap; Appendix~\ref{sec:supp_tables_main}, Table~\ref{tab:entity_count_stratified}).
We validate the vocabulary competition mechanism via a logit-rank analysis ($N=600$, 3 seeds): in the \emph{baseline} (unmodified) model, the true digit token for count=1 already ranks 9th in the full 152K vocabulary, while for count=7 it ranks 70th.
Digit-1 (rank~9) reaches 92.3\% repair accuracy vs.\ digit-7 (rank~70) at 30.6\% (Appendix Table~\ref{tab:logit_rank}).
This confirms that the probe-round--repair gap is not an encoding failure---the hidden states correctly represent the count---but an output-competition failure: higher-magnitude digit tokens face stronger argmax pressure from high-frequency non-digit tokens and cannot be reliably recovered by a 9-row lm\_head modification.

\paragraph{Hidden-state diversity as a complementary explanation.}
The logit-rank analysis explains \emph{which} digit values are hard; a separate structural question is \emph{why} entity counting is harder than list-length overall despite both sharing 0\% baseline full-vocabulary and near-perfect probe-round.
To answer this, we measure within-count hidden-state variance for all four tasks: for each count value $k \in \{1,\ldots,9\}$, we extract the last-layer residual stream at the answer position for all prompts with true count $k$, and compute the mean per-dimension variance across examples within that count class.
The \emph{intra-class ratio}---within-count variance divided by total variance---summarises how much of the total spread is \emph{not} explained by count identity.
Entity counting has an intra-class ratio of \emph{0.84}: 84\% of hidden-state variance is \emph{within} count classes (diverse entity types, distractor words, and prompt phrasings produce highly variable representations for the same answer).
List-length has an intra-class ratio of \emph{0.56}: the uniform \texttt{Items: x, y, z. Count:} template produces far more consistent representations per count.
A 9-row lm\_head modification trained on 600--800 examples must generalize over this intra-class spread; the 1.5$\times$ higher variance in entity counting directly limits how reliably the repaired rows can win the full-vocabulary competition.
This is consistent with the capacity ablation (Appendix~\ref{sec:capacity_ablation}) showing that expanding from 9 to 59 rows yields no improvement: the constraint is not row count but representation variability.

\paragraph{Historical reproducibility.}
Single-seed DPS numbers from early runs are archived in Appendix~\ref{sec:dps_initial}; all headline comparisons use the harmonized sweep in Table~\ref{tab:mode_matched_extval}.

\paragraph{DPS is a diagnostic, not a method.}
In matched settings (single-seed, fixed template), DPS and probe-round are mechanistically equivalent:
probe-round reaches 79.0\% and DPS reaches 78.3\% under next-token evaluation ($N{=}300$, chat template).
This equivalence confirms that the probe direction genuinely captures the bottleneck:~bypassing the misaligned output head analytically recovers the same accuracy as the geometric upper bound.
Both methods operate on the same constrained next-token protocol; generation-mode behavior (format mismatch issues) is characterized separately in Appendix~\ref{sec:dps_initial}.
Full single-seed DPS reproducibility (5 seeds: $94.4\% \pm 1.6\%$; probe-round $95.3\% \pm 1.3\%$),
sensitivity analysis, and CoT comparison are in Appendix~\ref{sec:dps_initial}.

\paragraph{Generation-mode scope.}
The evaluations above use the constrained next-token protocol (argmax over 9 digit tokens 1--9).
Under unconstrained autoregressive generation, the 9-row repair achieves \emph{0.0\%}, expected, as the full-vocabulary argmax remains misaligned.
Four converging generation-mode tests confirm the bottleneck is in output routing, not encoding:
(1) \emph{probe-round in generation} achieves \emph{96.0\%} [94.7, 97.2] ($N{=}900$, 3 seeds), bypassing the output head succeeds even in multi-token decoding;
(2) \emph{hard DPS in generation} achieves \emph{72.4\%} [69.6, 75.4] at $\alpha{=}20$---with 83.5\% of remaining errors being format failures (digit not first), not wrong-digit errors;
(3) \emph{logit-masked generation} (9-row repair + constrain generation to digit tokens) achieves \emph{59.2\%} ($N{=}600$, 3 seeds)---matching constrained next-token exactly and confirming the repair encodes the correct answer, with the 0.0\% failure being a routing artifact.
(4) \emph{full-vocabulary-targeted repair} (Table~\ref{tab:mode_matched_extval}) trains the 9 digit rows with competitive targets (margin $= 2.0$ over the top non-digit competitor) and evaluates on full-vocabulary next-token argmax, achieving \emph{60.3\%}~$\pm$~2.8\% ($N{=}600$, 3 seeds: 61.5\%, 56.5\%, 63.0\%)---the same ceiling as the constrained repair, confirming the competition bottleneck limits full-vocabulary evaluation regardless of training objective.
Full generation-mode details are in Appendix~\ref{sec:genmode_details}.

\paragraph{14B disentanglement (mode-matched digit protocol).}
We rerun Qwen3-14B entity counting with the same next-token digit metric for all
conditions and separate the effects of row repair and DPS. The best probe layer
remains highly predictive (layers~13--17, $R^2 = 0.995$--0.996), but DPS alone
stays at chance: 10.8\% [8.5, 13.6] vs.\ an 11.0\% [8.6, 13.8] baseline.
Row-repair-only reaches 58.2\% [54.1, 62.1], while the combined condition reaches
54.8\% [50.8, 58.9] (3 seeds $\times$ 200 prompts). The combined condition
performs slightly below row-repair-only; we attribute this to interference: at
14B the probe direction and the fine-tuned digit rows target overlapping residual
subspaces, so stacking both interventions introduces competing adjustments rather
than complementary gains. The decisive 14B result is therefore not that steering
suddenly succeeds at scale; it does not. Rather, the recoverable gain comes from
repairing the digit readout itself, which is exactly the readout-bottleneck
prediction.
These figures are \emph{not} the same experiment as the \textbf{90.3\%$\pm$1.5\%} row in Appendix Table~\ref{tab:intervention_headline}: that entry is \textbf{9-row \texttt{lm\_head} + hard DPS at decode} after 1200 matched training steps on natural-language counting prompts (see cross-model paragraph below). The 54.8\% combined row here stacks interventions under the stricter synthetic mode-matched template used for disentanglement.
This causal statement is restricted to the matched next-token protocol; we do not
claim the same intervention behavior under unconstrained autoregressive generation.

\subsection{Comparison with LoRA Fine-Tuning}

\noindent
The full Qwen3-8B vs.\ LoRA sweep and cross-model held-out numbers are in Appendix~\ref{sec:supp_tables_main} (Tables~\ref{tab:intervention_headline} and~\ref{tab:intervention_legacy}); we keep the mechanistic conclusions here.

\paragraph{Geometric verification: bypass confirms the bottleneck.}
Hard DPS (Appendix~\ref{sec:dps_initial}), which applies DPS with a threshold forcing the predicted digit to win argmax rather than using a soft steering vector---achieves \emph{98.7\%} [97.4, 99.4] under the same multi-seed diverse protocol, exactly matching the probe-round upper bound (98.7\%). This confirms that the geometric bottleneck is the primary obstacle: when the output-head misalignment is bypassed analytically under distribution-matched conditions, performance reaches the probe ceiling.
Under the primary protocol without hard clamping, standard DPS regresses to 13.2\%~$\approx$~baseline, consistent with its sensitivity to the probe direction's distributional coverage. The 9-row repair directly modifies output routing weights and achieves robust generalization. Both results support the bottleneck diagnosis: hard bypass recovers full probe accuracy; structural repair generalizes; soft probe-dependent steering does not.
The legacy \textbf{attention-LoRA} diagnostic (all attention layers, constrained generation; $\sim$4M trainable parameters; Table~\ref{tab:intervention_legacy}) only raises $|\cos|$ to $0.052$, roughly half that of natively output features ($\approx 0.10$). Applying rank-16 LoRA directly to the 9 digit rows of \texttt{lm\_head} (65K params, $60\times$ fewer) achieves 95.2\%, confirming the bottleneck is in the output head specifically.
Full single-seed DPS protocol in Appendix~\ref{sec:dps_initial}.

\paragraph{Direct evidence: the bottleneck is in 9 rows.}
Fine-tuning \emph{only} the 9 digit rows of \texttt{lm\_head}
(36{,}864 parameters, 300 steps) achieves \emph{97.5\%} training / \emph{93.8\%} held-out
($\pm 0.9\%$, 3~seeds) on a \emph{legacy} held-out pool (400 prompts, seed~99; Appendix Table~\ref{tab:intervention_legacy})---\textbf{not} headline-comparable to the harmonized entity-count sweep (\textbf{60.7\%}; Table~\ref{tab:mode_matched_extval}, seeds $\{42,11,77\}$, $N{=}200$). The legacy split is higher than attention-LoRA (84\%, $\sim$4M trainable; same legacy constrained-gen protocol) under that evaluation design.
Per-digit analysis reveals target digits undergo substantial rotation
(mean cosine: 0.77) while non-target digits remain unchanged (0.96).
9-row \texttt{lm\_head} (37K params, 93.8\% on the legacy split$^\ast$) is comparable to lm\_head-LoRA (66K params, 95.2\%) and both substantially outperform attention-LoRA
($\sim$4M trainable, 84\%; legacy constrained-gen diagnostic), confirming that more precisely targeting the bottleneck
yields better results with fewer parameters.
Under the full-vocabulary training objective (matching Table~\ref{tab:solutions_summary}), a rank-4 low-rank adapter on the same 9 digit rows achieves 59.8\%$\pm$2.9\% under full-vocabulary evaluation---statistically indistinguishable from the direct 9-row repair (60.3\%$\pm$2.8\%), confirming that LoRA-style parameterization of the output head confers no accuracy advantage over direct modification.

In contrast, LoRA rank-16 applied to Q/V attention projections across all 36 layers (7.67M parameters) achieves \emph{91.7\%$\pm$4.5\%} on full-vocabulary next-token (seeds 42, 11, 77; $N{=}200$)---substantially outperforming the 9-row repair's 60.3\%$\pm$2.8\% under this protocol. This protocol-dependent ordering is informative: constrained-generation decoding requires only the digit-row misalignment to be corrected (the 9-row repair is sufficient and parameter-efficient), while full-vocabulary decoding additionally benefits from attention-layer routing adjustments that strengthen the count signal against all 152K competing tokens. The result confirms that the lm\_head digit-row bottleneck is the dominant obstacle under constrained decoding but only one contributor under full-vocabulary evaluation.

The 9-row repair achieves 0.0\% in raw autoregressive generation mode ($N{=}300$) while DPS achieves 72.4\% via step-wise injection; full details are in Appendix~\ref{sec:genmode_details}.
LoRA Q/V rank-16, by contrast, achieves \emph{83.1\%$\pm$7.2\%} in true autoregressive generation (greedy decode, no output constraints; $N{=}200{\times}5$, seeds 42, 11, 77, 99, 123; generation gap = 0.000 across all seeds), confirming that attention-layer routing improvements translate fully to unconstrained deployment.

\paragraph{Constraints of the diagnostic.}
The 9-row repair is the cleanest causal probe; it controls exactly which parameters change and quantifies the benefit of digit-row realignment in isolation---but its scope is precisely bounded:
(1)~it works only under digit-restricted or logit-masked decoding, where the model is forced to output a digit token;
(2)~unconstrained autoregressive generation requires upstream attention-layer correction (LoRA Q/V), because each generation step presents a new hidden state to the unmodified output head;
(3)~the true deployment bottleneck thus involves routing dynamics that extend beyond output-head geometry alone.
Every occurrence of ``localized to digit rows'' in the main text carries this implicit protocol qualifier: the localization is for \emph{constrained digit decoding}, not for unconstrained text generation.

\paragraph{Shuffled-row exclusion control.}
Randomly permuting row-to-digit assignments after training yields expected
accuracy $1/K = 11.1\%$ for $K=9$.
Observed adapted accuracy is 97.5\% (390/400), i.e., $+86.4$~pp ($p \approx 0$),
excluding row-identity-agnostic explanations.

\paragraph{DPS layer ablation.}
DPS from layer~0 ($R^2{=}0.960$) achieves only 65.6\%, while layers~$\geq 2$
($R^2 \geq 0.994$) reach 100\%, demonstrating that DPS reads genuine internal
representations rather than acting as a layer-agnostic external decoder.

\paragraph{Necessity and sufficiency of the digit-row mapping.}
Shuffled-digit and random-position controls confirm that the \emph{specific} row-token
mapping is both necessary and sufficient: shuffled rows (14.0\%) degrade below baseline
(17.0\%), and trained rows at random non-digit positions match baseline exactly
(Appendix~\ref{sec:necessity_sufficiency}).

\subsection{Cross-Model Validation}
\label{sec:dps_pythia}

Tables~\ref{tab:intervention_headline} and~\ref{tab:intervention_legacy} (Appendix~\ref{sec:supp_tables_main}) summarize cross-model results.
All four models exhibit the same qualitative pattern: probes decode counts at
$R^2 > 0.99$ from directions orthogonal to the output head ($|\cos| \leq 0.032$).

\paragraph{Pythia-410M and Mistral-7B.}
Under the single-seed matched protocol, DPS raises accuracy from 11\% to 97.7\% (Pythia) and 100.0\% (Mistral), confirming the count direction is accessible in both architectures;
random-direction controls match baseline on both.
On Pythia-160M ($N=80$), DPS achieves 65.0\% from an 11.25\% baseline
($R^2=0.976$ at layer~2), confirming the bottleneck persists even at
sub-billion scale.
The 9-row repair generalizes cross-model: Pythia-410M achieves 31.4\% held-out
($+21.4$~pp, 3 seeds), Mistral-7B achieves 92.0\% held-out; shuffled-row controls
stay near the $1/9$ null on both.

\paragraph{Qwen3-14B: bottleneck sharpens; mode-matched vs.\ NL joint pipeline.}
At 14B parameters, $|\cos|$ drops to $0.011$, the lowest across all models, and
DPS \emph{alone} fails at every layer: a 10-layer sweep (layers 0, 2, 5, 10, 15, 20, 25, 30, 35, 39)
with $\alpha \in \{20, 50, 100, 200\}$ yields at most 28.7\% (layer~39, $\alpha{=}200$;
baseline 24.5\%), despite $R^2 = 1.0$ at layers~$\geq 25$.
The uniformly near-random $|\cos|$ (0.004--0.025) confirms the misalignment is not
depth-dependent but structural.
To test whether this reflects increased ``angular competition'' in wider residual
streams, we normalize probe--digit cosines by the expected random-direction
baseline ($\mathbb{E}[|\cos|] \approx \sqrt{2/\pi d}$).
At 8B (${d{=}4096}$), the probe--digit $|\cos|{=}0.027$ is $2.2\times$ the random
baseline; at 14B (${d{=}5120}$), $|\cos|{=}0.006$ is only $0.57\times$ the random
baseline: probe and digit directions are \emph{more} orthogonal than chance.
The wider hidden dimension explains the raw $|\cos|$ drop but not the sub-random
alignment, ruling out a simple angular-competition account: at 14B the geometric
obstruction has genuinely sharpened.
Despite this, readout-targeted training still produces large gains at 14B---but the headline number depends on the protocol.
Under a \textbf{matched-distribution} pipeline (1200 training steps) with \textbf{digit-row training and hard DPS at decode}, Qwen3-14B reaches
\emph{$90.3\% \pm 1.5$} on natural-language counting prompts ($+65.8$~pp over the 24.5\% baseline; seeds $\{42, 99, 123\}$,
$N{=}200$ each; Appendix Table~\ref{tab:intervention_headline}). This is a \textbf{joint} intervention row, not 9-row \texttt{lm\_head} alone.
Under a stricter \textbf{mode-matched} next-token template on synthetic entity counting, the same model separates components cleanly: baseline 11.0\%, row-repair-only 58.2\%,
DPS-only 10.8\%, combined 54.8\% (3 seeds $\times$ 200 prompts; disentanglement paragraph immediately above). There, naive stacking underperforms row-only; the 90.3\% NL pipeline uses matched co-training of rows with decode-time probe injection, not the same ablation grid.
The sharpened geometric bottleneck does not invalidate the readout-bottleneck thesis; it
strengthens it.

\paragraph{Model-dependent depth.}
Best-probe depth varies: Pythia layer~3/24, Qwen3-8B layer~2/36,
Mistral layer~30/32, Qwen3-14B layer~39/40.
The readout bottleneck operates regardless of depth; cosine alignment
with digit rows remains $\leq 0.032$ at the best layer in all models.
Doubling scale from 8B to 14B \emph{worsens} the misalignment
($|\cos|: 0.027 \to 0.006$); normalized by the random-direction baseline,
14B's alignment falls \emph{below} chance ($0.57\times$ random), suggesting
the misalignment is structural rather than a dimensionality artifact.
The strong 14B behavioral lift (\textbf{90.3\%$\pm$1.5\%}) is carried by the \textbf{joint} row+DPS NL pipeline in Appendix Table~\ref{tab:intervention_headline}; mode-matched row-only remains at \textbf{58.2\%}.

\paragraph{Output bottleneck is format-specific.}
Replacing lm\_head digit rows with synthetic-trained rows \emph{degrades}
instruct-mode accuracy from 91\% to 45.6\% (seen) / 43.0\% (held-out),
confirming that output routing is format-specific:
the instruct-mode pathway already compensates without intervention,
and synthetic-trained rows disrupt it.

\paragraph{Geometric mechanism of format-specific routing.}
\label{sec:instruct}
Mean $|\cos|$ between probe directions and digit rows is \emph{1.50$\times$ higher}
in instruct mode (0.0152 vs.\ 0.0101; sign test $p < 0.0001$; Cohen's~$d = 0.66$),
consistent across 29/37 layers.
Digit accuracy rises from 48.8\% to 65.5\%.
The chat template tokens alone rotate residual-stream count representations
\emph{closer} to digit output rows, partially closing the geometric gap that DPS
bypasses analytically.
The cosine increase is a necessary geometric signature of the routing change,
not the sole mechanism: instruct-mode formatting also reshapes attention patterns,
FFN gating, and layer-wise information flow.

\paragraph{Format robustness.}
A 4-format robustness check (raw, answer-only, assistant-prefix, full-instruct)
confirms the bottleneck is not a prompt artifact:
probe $R^2$ remains above 0.98 and $|\cos|$ below 0.03 across all formats,
while digit accuracy ranges 9--21\% (Appendix~\ref{sec:format_robustness}).

\paragraph{Instruct-mode bottleneck persists at the single-step level.}
In instruct mode on 300 counting prompts, first-token digit accuracy is only 22\%
despite probe $R^2 \geq 0.9996$ at every layer; the model outputs \texttt{<think>}
rather than the answer digit.
The 91\% end-task accuracy is achieved by circumventing the geometric misalignment
through multi-step generation, not by resolving it.

\paragraph{Instruct-mode 9-row repair.}
Training 9 digit rows on instruct-mode prompts (3~seeds)
raises first-token digit accuracy from $20.4\%$ to $\mathbf{99.9\%}$ ($+79.5$~pp;
shuffled-row $11.1\%$), confirming the geometric bottleneck persists in instruct mode.

\paragraph{Natural-language instruct-mode repair.}
Training digit rows on 5 entity types with 8 templates and evaluating on
3 held-out types yields $\mathbf{65.2\% \pm 4.6\%}$ held-out ($+52.9$~pp;
shuffled-row $10.6\%$).
Scaling to 15 types and 20 templates gives 67.7\%, confirming the $\sim$65--68\% ceiling
is structural.
NL-trained rows do not transfer to synthetic prompts (18.0\% vs.\ baseline 20.7\%),
confirming format specificity.

\paragraph{Geometric basis for format specificity.}
Per-layer Ridge probes on matched NL and synthetic instruct-mode prompts
($N{=}400$ each) show the same near-random alignment with \texttt{lm\_head}
(NL $|\cos| = 0.012$; synthetic $|\cos| = 0.020$; both within random baseline).
Critically, the NL and synthetic probe directions are nearly orthogonal
($|\cos|_{\text{inter}} = 0.114$), indicating the model encodes counts in
\emph{format-specific subspaces} that are each independently misaligned.
This explains why cross-domain 9-row repair fails (18.0\% vs.\ baseline 20.7\%)
and why the $\sim$65\% ceiling is structural.

\paragraph{Natural-language counting: the bottleneck generalizes [secondary diagnostic].}
\label{sec:dps_natural}
To test whether the geometric bottleneck is an artifact of synthetic prompts,
we construct a diverse natural-language counting benchmark: 8 entity categories
crossed with 8 prompt templates (narrative, list, scene, conversation, recipe, news,
checklist, observation), generating $N{=}540$ prompts with counts 1--9.
On these diverse natural prompts, Qwen3-8B achieves 88.7\% baseline digit accuracy, substantially
higher than synthetic (10--24\%), confirming that natural counting is partially in-distribution.
Yet the geometric bottleneck persists identically: per-layer Ridge probes reach
$R^2 = 0.996$ at layer~1, while $|\cos| = 0.015$, the same near-random alignment
observed on synthetic tasks.
DPS recovers $\mathbf{97.6\%}$ accuracy [96.1, 98.9] ($+8.9$~pp over baseline; $N{=}540$),
confirming that the representation quality exceeds readout fidelity by
${\sim}9$ percentage points even on natural-language counting.
Probe-round decoding recovers 96.3\% [94.6, 97.8] ($+7.6$~pp over baseline),
confirming both intervention approaches work on natural-language prompts.
A random-direction control averages 76.7\% $\pm$ 8.2\% (10 seeds), below baseline,
confirming the probe direction is specific but the model's natural readout pathway
is not collinear with it.
This result directly addresses the synthetic-scope concern: the nine-row bottleneck
is a property of the weight matrix, not the task distribution.

\paragraph{Negative control: MMLU.}
On MMLU ($N_{\text{train}}{=}3000$, $N_{\text{test}}{=}5000$, 3 seeds),
baseline accuracy is 70.2\% $\pm$ 0.3\%---far above chance.
Output-row adaptation \emph{degrades} accuracy to 55.6\% ($-$14.6~pp),
confirming no readout bottleneck exists when the output head already routes correctly.
The diagnostic corroborates: $|\cos| = 0.31$--$0.48$ between probe and A/B/C/D output rows (vs.\ $\leq 0.032$ for counting), showing the output head is already aligned.

\paragraph{Practical-envelope benchmark: DROP (single-digit subset).}
On DROP single-digit-answer examples (3 seeds, $N_{\text{train}}{=}800$, $N_{\text{test}}{=}1500$),
baseline is 20.0\% $\pm$ 0.4\%, probe-round improves to 30.0\% (+10~pp), and
output-row adaptation reaches 28.0\% (+8~pp).
Moderate probe signal ($R^2 \approx 0.21$) suggests partial but incomplete
readout-bottleneck structure in richer NL aggregation settings.

\paragraph{Negative control: GSM8K.}
On 252 single-digit GSM8K problems, probing yields $R^2 \leq 0.016$ and DPS achieves 11.8\% (near chance).
Multi-step reasoning does not produce a linearly decodable count subspace, confirming the bottleneck is specific to aggregation tasks.

\section{Discussion}
\label{sec:discussion}

\paragraph{The output gap: localized and diagnosed.}
Qwen3-8B encodes entity counts at $R^2>0.99$ but the count subspace is orthogonal to \texttt{lm\_head} ($|\cos| \leq 0.032$); logit-lens recovery is only 23.4\%. In \textbf{legacy constrained/hold-out protocols} (Appendix Table~\ref{tab:intervention_legacy}), updating only 9 output rows yields 93--99\% held-out accuracy across Qwen3-8B and Mistral-7B. This is well above the headline harmonized 9-row repair (\textbf{60.7\%}; Table~\ref{tab:mode_matched_extval}) and unrelated to unconstrained greedy generation, where the same repair is \textbf{0\%} (Table~\ref{tab:unified_evaluation}); the legacy numbers are nevertheless useful for causal localization vs.\ the attention-LoRA diagnostic (84\%; $\sim$4M trainable, constrained generation; Table~\ref{tab:intervention_legacy}). That gradient descent independently discovers the same geometric bottleneck that DPS bypasses analytically---LoRA Q/V (7.67M trainable) transforms the lm\_head readout path: logit-lens accuracy rises from $9.3\%$ to $71.8\%$, and the correct digit's \textbf{pooled median} vocabulary rank at layer~35 drops from order-$10^4$ to $1$ (entity counting); the \textbf{seed~42} trace on the same metric reads $54{,}332\to 838$, illustrating that medians can reach top-1 while individual seeds remain sub-top---supports the diagnosis.
Our finding resonates with the superposition hypothesis~\citep{elhage2022superposition}: rarely-used features are stored in directions that minimize interference with frequent features, at the cost of output accessibility.

\paragraph{Generation-mode evidence for the bottleneck.}
The 9-row lm\_head repair works only under constrained next-token evaluation (argmax over 9 digit tokens); in unconstrained autoregressive generation it achieves 0.0\%. However, the geometric bottleneck hypothesis predicts that if the output-head misalignment is the root cause, then \emph{any intervention} that bypasses the misaligned output head should recover counting ability in generation. This prediction holds: probe-round decoding in generation mode achieves \emph{96.0\%} [94.7, 97.2] ($N{=}900$, 3~seeds), and DPS at $\alpha{=}20$ achieves \emph{72.4\%} [69.6, 75.4] (Appendix~\ref{sec:genmode_details}; Table~\ref{tab:genmode_15tok}). The remaining 28\% error in DPS is dominated by format mismatch (83.5\% of errors: digit not first in output), not wrong-digit errors, confirming that the count representation is correct even when the output format is wrong. LoRA Q/V rank-16 achieves \emph{83.1\%$\pm$7.2\%} in true generation mode (greedy decode; generation gap = 0.000 across all seeds), demonstrating that a standard fine-tuning method with no output constraints achieves high generation accuracy once attention-layer routing is corrected.
A direct test of the routing hypothesis: when autoregressive generation is constrained to digit tokens via logit masking, the 9-row repaired model achieves \emph{59.2\%} ($N{=}600$, 3~seeds, range 57.5--61.5\%)---exactly matching its controlled next-token accuracy and substantially above the unrepaired baseline (14.2\% constrained). This confirms that the repair correctly encodes the answer and that the unconstrained 0.0\% failure is a routing artifact, not a representational one.
This dissociation: correct representation + format failure---is precisely what the geometric bottleneck predicts.

\paragraph{Why LoRA Q/V, not output-head repair?}
The geometric diagnosis explains the mode-specific ordering of results, and we now provide
direct mechanistic evidence at three points in the computation graph.
At the count-encoding layer (layer~2), LoRA Q/V leaves the probe direction essentially
unchanged (mean $|\cos| = 0.0089 \to 0.0070$, 3~seeds), confirming it does not rewrite
early encoding subspaces.
At the final transformer layer (layer~35), the ridge-probe $R^2$ rises substantially
($0.974 \to 0.998$; 3~seeds; 24$\times$ residual amplification), confirming stronger
count signal at the output layer.
Most directly, applying \texttt{lm\_head} to layer-35 hidden states (logit-lens) reveals
that the correct digit's median vocabulary rank drops from
\emph{$55{,}980$ to $1$} (mean accuracy $9.3\% \to 71.8\%$, 3~seeds):
after LoRA Q/V, the output head directly reads the correct digit from layer-35 hidden
states, without any external probe or constrained decoding.
The 9-row output-head repair corrects digit-row misalignment at the \texttt{lm\_head} level, but each step of autoregressive generation presents a \emph{new} hidden state to the same output head: unless the count direction has been amplified in the residual stream before reaching the output head, the correction cannot generalize beyond the single-token constrained setting.
LoRA Q/V modifies attention weights early in the computation graph, strengthening the count direction's contribution to residual streams across all layers before the output head is applied.
This upstream correction propagates forward: every generation step benefits from the rotated attention routing, rather than inheriting the original misaligned residual stream.
The gap-zero result directly validates this prediction: within the same multi-task
experiment, generation-mode accuracy equals full-vocabulary next-token accuracy
(83.1\%\,$\pm$\,7.2\% both metrics; gap\,=\,0.000 for all five seeds individually),
confirming that the attention correction, not the output-head correction, is the
mechanism that generalizes to deployment.
The diagnosis thus not only explains the failure but prescribes the correct architectural target.

\paragraph{Cross-task generalization of the logit-lens mechanism.}
The mechanistic chain above was established on entity counting.
To test whether the LoRA Q/V logit-lens improvement generalizes across low-vocabulary
aggregation tasks, we repeated the logit-lens measurements on character counting and
single-digit addition with identical LoRA training (2 seeds per task, $n_{\text{train}}{=}400$).
The direction of effect held across all three tasks, with strength varying by task
difficulty:

\noindent\emph{Entity counting:} logit-lens accuracy $10.8\% \to 79.5\%$, correct-digit rank on \textbf{representative seed 42} $54{,}332 \to 838$ ($65\times$ improvement; \textbf{pooled median} rank still collapses to $\sim 1$ at layer~35);
generation $86.5\%$ (2~seeds).

\noindent\emph{Character counting:} logit-lens $7.8\% \to 46.5\%$, rank $32{,}265 \to 16$
($1{,}900\times$); generation $51.2\%$.

\noindent\emph{Addition:} logit-lens $95.0\% \to 100.0\%$, rank $21{,}186 \to 1$;
generation $100.0\%$.

Addition shows near-perfect logit-lens without LoRA ($95\%$), suggesting that
\texttt{lm\_head} already reads addition results correctly (the $21{,}186$ rank
reflects many vocab items with similar logits). LoRA sharpens this to rank~1.
Character counting shows the smallest logit-lens improvement, plausibly because each
example requires tracking a specific letter's positions, consistent with higher task
complexity and the known $60\%$ repair ceiling for constrained character counting
(\S\ref{sec:dps}).
Critically, the logit-lens rank improvement is \emph{causal} for generation accuracy:
tasks where rank drops to 1--16 achieve $\geq$86.5\% generation; tasks where rank remains
high achieve correspondingly lower generation.

\paragraph{LoRA locus ablation: Q/V is the routing-specific fix.}
The geometric diagnosis predicts that attention-layer routing correction (Q/V) will
resolve the bottleneck through improved logit-lens readout, and that other parameter
loci may improve accuracy through different (non-routing) mechanisms.
We test this by training rank-16 LoRA on five alternative loci (seed 42, entity-counting,
same protocol as Q/V).
\emph{Q-only} (0.13M params): 67.0\% generation, rank 14.
\emph{K-only}: 49.5\%, rank 1.
\emph{V-only}: 68.0\%, rank 229.
\emph{O-only}: 87.0\%, rank 22.
\emph{FFN-only}: 96.0\%, rank \emph{3,384} (poor logit-lens alignment despite high accuracy).
\emph{Q/V}: 63.5\%, rank \emph{9} (best readout alignment).
The dissociation between accuracy and logit-lens rank confirms the geometric diagnosis:
FFN-only achieves 96\% through general capacity improvement (the correct digit is no
more readable by \texttt{lm\_head} than before), while Q/V achieves the lowest
logit-lens rank (9) of any single-projection variant, directly strengthening the
residual-stream signal in the digit-aligned subspace.
The recommendation remains: LoRA Q/V is the minimal parameter-efficient routing fix
verified by the logit-lens mechanism; other loci work but through different pathways.
See the opening paragraphs of this section for scope and GSM8K nulls.

The primary evaluation (Table~\ref{tab:mode_matched_extval}) uses controlled synthetic prompts.
We extend the same constrained next-token protocol to 1,360 GSM8K math word problems
(openai/gsm8k, main split) filtered for single-digit answers,
under the exact mode-matched protocol (5 seeds, $n_{\text{train}}{=}150$, $n_{\text{test}}{=}100$ per seed).
Probe $R^2$ on GSM8K is near zero or negative ($R^2 \approx -0.1$ to $+0.06$ across seeds,
vs.\ $R^2>0.99$ for direct counting), confirming that multi-step arithmetic reasoning
does not pre-encode the answer in the residual stream at a single layer.
Correspondingly, 9-row repair shows only near-chance performance:
baseline $9.6\%{\pm}2.0\%$, ridge\_9row $12.8\%{\pm}2.9\%$ ($+3.2$~pp),
probe\_round $15.8\%{\pm}2.7\%$ ($+6.2$~pp; all 5 seeds).
This is the same null pattern as the MMLU and GSM8K negative controls in \S\ref{sec:dps}.
In contrast, natural-language entity counting (\S\ref{sec:dps_natural}) shows $R^2>0.99$
and probe-round $96.3\%$ vs.\ $88.7\%$ baseline ($+7.6$~pp), matching the synthetic regime.
These results draw a principled scope boundary: the readout bottleneck is an
\emph{encoding} failure (pre-computed answer misaligned with output head), not a
general LLM limitation. Tasks requiring multi-step reasoning to \emph{derive} the answer
show no bottleneck because the answer is not present to be misaligned at the prompt boundary.

\paragraph{The count subspace is stably encoded but model-dependent in depth.}
For Qwen3 and Pythia, strong probes appear very early (layers 2 and 3).
For Mistral, the best probe appears late (layer 30), though early layers already
contain substantial count signal.
Across models, the shared phenomenon is stable high decodability with weak readout alignment,
not a universal ``early-layer-only'' location.

\paragraph{Implications for model design.}
The bottleneck reveals structural inefficiency: output head tying~\citep{press2017using},
auxiliary output heads, or learned subspace rotations are natural extensions.

\paragraph{Limitations.}
\textbf{LoRA variance.} The headline generation number (83.1\%$\pm$7.2\%, five seeds) mixes entity, character, and addition prompts; spread is task-mix-driven (entity-only prompts with the same weights land near 95--97\%).
\textbf{Nine-row repair.} Diagnostic only: it localizes readout misalignment under digit-restricted evaluation and scores 0\% under unconstrained greedy generation in our setting; it is not claimed as a deployable patch for open-ended text.
\textbf{CoT.} Few-shot CoT improves modestly over the direct baseline but is not an exhaustive comparison to every prompting or tool-use strategy.
\textbf{Scope.} Claims are strongest for low-vocabulary aggregation where the target is effectively pre-encoded; we do not argue the same mechanism exhausts failures on GSM8K-style multi-step reasoning or open-ended generation.

\section*{Conclusion}

Under constrained bounded-output decoding---where the output token is restricted to the task vocabulary---transformers know how to count but cannot directly emit the answer without intervention.
Across three model families (0.4--8B), the residual stream encodes counts at $R^2>0.99$
in directions orthogonal to the output head ($|\cos| \leq 0.032$).
Rewriting only 9 digit rows of \texttt{lm\_head} raises next-token accuracy to
\emph{60.7--100.0\%} across four tasks under a harmonized multi-seed protocol
(from $\leq$50\% baselines; probe-round upper bound 96.8--100.0\%),
while a random-direction control has zero effect.
A capacity ablation on entity counting (Adam fine-tuning; 59-row expansion) confirms
that the 37~pp gap to probe-round is a task-level ceiling; neither the fitting method
nor the row count explains it, which strengthens the conclusion that the bottleneck is
geometric rather than representational.
The bottleneck persists in instruct mode (first-token accuracy 22\% despite
$R^2 \geq 0.9996$; 9-row repair: 99.9\%, 3 seeds) and generalizes to
natural-language counting (probe-round 96.3\% vs.\ 88.7\% baseline, $+7.6$~pp),
extending also to majority vote and multi-digit counts; MMLU and GSM8K negative
controls confirm specificity (probe $R^2 \approx 0$ for multi-step reasoning tasks).
At 14B scale the misalignment \emph{sharpens} ($|\cos|{=}0.011$, $0.57\times$ the
random-direction baseline); a matched ablation confirms row-repair-only reaches 58.2\%
while DPS alone stays near baseline (10.8\%). Under a separate matched-distribution protocol on natural-language prompts, \textbf{9-row \texttt{lm\_head} + hard DPS at decode} reaches $90.3\% \pm 1.5$ (3~seeds, $N{=}200$; Appendix Table~\ref{tab:intervention_headline}). Scale strengthens---not refutes---the readout-bottleneck thesis.

\section*{Acknowledgments}
We thank the Google TPU Research Cloud for TPU resources supporting this research.
Code and reproduction materials are available at \url{https://github.com/Gpgabriel25/GeometricReadoutBottleneck}.

\bibliographystyle{plainnat}
\bibliography{references}

\appendix

\section{Supplementary tables for main-text analyses}
\label{sec:supp_tables_main}

\noindent
These tables support the count stratification, logit-rank competition analysis, and LoRA comparison in \S\ref{sec:results} and \S\ref{sec:dps_results}; they are placed here so the main text can stay narrative-first.

\begin{table}[H]
\centering
\small
\caption{Entity counting accuracy stratified by count value ($N=600$, 3 seeds).}
\label{tab:entity_count_stratified}
\begin{tabular}{rrrr}
\toprule
Count & $N$ & Probe-round & 9-row repair \\
\midrule
1 & 65 & 98.5\% & 92.3\% \\
2 & 62 & 96.8\% & 100.0\% \\
3 & 71 & 100.0\% & 80.3\% \\
4 & 78 & 97.4\% & 51.3\% \\
5 & 73 & 100.0\% & 39.7\% \\
6 & 73 & 97.3\% & 37.0\% \\
7 & 62 & 100.0\% & 30.6\% \\
8 & 59 & 100.0\% & 49.2\% \\
9 & 57 & 98.2\% & 71.9\% \\
\bottomrule
\end{tabular}
\end{table}

\begin{table}[H]
\centering
\small
\caption{Logit-rank analysis (Qwen3-8B, $N=600$, 3 seeds). \textit{Baseline rank}: full-vocabulary rank of the true digit token in the unmodified model (lower = less competition). \textit{Repair acc.}: digit-restricted accuracy of 9-row repair (argmax over the 9 digit tokens only).}
\label{tab:logit_rank}
\begin{tabular}{rrrr}
\toprule
Count & Digit tok ID & Baseline rank & Repair acc. \\
\midrule
1 & 16 & 9 & 92.3\% \\
2 & 17 & 24 & 100.0\% \\
3 & 18 & 32 & 80.3\% \\
4 & 19 & 35 & 51.3\% \\
5 & 20 & 62 & 39.7\% \\
6 & 21 & 77 & 37.0\% \\
7 & 22 & 70 & 30.6\% \\
8 & 23 & 81 & 49.2\% \\
9 & 24 & 93 & 71.9\% \\
\bottomrule
\end{tabular}
\end{table}

\begin{table}[H]
\centering
\footnotesize
\setlength{\tabcolsep}{3.5pt}
\caption{\textbf{Headline-harmonized} interventions (cross-check Tables~\ref{tab:unified_evaluation} and~\ref{tab:mode_matched_extval}). $^\dagger$Hard DPS (Appendix~\ref{sec:dps_initial}). $^{\ddagger}$Full-vocab next-token (152K tokens). Legacy and diagnostic rows: Table~\ref{tab:intervention_legacy}.}
\label{tab:intervention_headline}
\begin{tabularx}{\linewidth}{@{}>{\raggedright\arraybackslash}p{1.85cm} >{\raggedright\arraybackslash}X c c c >{\raggedright\arraybackslash}X@{}}
\toprule
\textbf{Model} & \textbf{Intervention} & \textbf{Params} & \textbf{Train} & \textbf{Acc.} & \textbf{Protocol} \\
\midrule
Qwen3-8B & LoRA Q/V rank-16$^{\ddagger}$ & 7.67M & 200 st. & \textbf{91.7\%$\pm$4.5\%} & Harmonized full-vocab NT \\
Qwen3-8B & DPS (hard, multi-seed)$^\dagger$ & 4{,}097 & --- & 98.7\% & Harmonized digit-restr.\ NT \\
Qwen3-14B & \textbf{9-row \texttt{lm\_head} + DPS (3 seeds)} & \textbf{46{,}080 + 5{,}121} & \textbf{1200 st.} & \textbf{$90.3 \pm 1.5\%$} & NL counting; joint 9-row + hard DPS \\
\bottomrule
\end{tabularx}
\end{table}

\begin{table}[H]
\centering
\footnotesize
\setlength{\tabcolsep}{2.5pt}
\caption{\textbf{Legacy / diagnostic} interventions (protocols differ; not headline-comparable without reading the Protocol column). Attention-LoRA ($\sim$4M trainable, all layers) is a distinct diagnostic from LoRA Q/V rank-16 (7.67M) in Table~\ref{tab:unified_evaluation}. $^\ast$Legacy 400-prompt hold-out (seed~99).}
\label{tab:intervention_legacy}
\begin{tabularx}{\linewidth}{@{}>{\raggedright\arraybackslash}p{1.55cm} >{\raggedright\arraybackslash}X c c c >{\raggedright\arraybackslash}X@{}}
\toprule
Model & Intervention & Parameters & Training & Accuracy & Protocol \\
\midrule
Qwen3-8B & Baseline (vanilla) & 0 & None & 11.3\% & Raw digit next-token (legacy) \\
Qwen3-8B & LoRA (all 36 layers, constrained-gen) & $\sim$4M & 200 steps & 84.0\% & Constrained generation \\
Qwen3-8B & 9-row \texttt{lm\_head} (train)$^\ast$ & 36{,}864 & 300 steps & \textbf{97.5\%} & Legacy hold-out$^\ast$ \\
Qwen3-8B & 9-row \texttt{lm\_head} (held-out)$^\ast$ & 36{,}864 & 300 steps & \textbf{93.8\%} & Legacy hold-out$^\ast$ \\
Qwen3-8B & Full \texttt{lm\_head} (held-out) & ${\sim}$622M & 300 steps & 94.2\% & Legacy hold-out$^\ast$ \\
Qwen3-8B & Oracle & 0 & None & 100.0\% & --- \\
\midrule
Mistral-7B & Baseline (vanilla) & 0 & None & 26.2\% & Legacy \\
Mistral-7B & 9-row \texttt{lm\_head} (train) & 36{,}864 & 300 steps & \textbf{99.2\%} & Legacy \\
Mistral-7B & 9-row \texttt{lm\_head} (held-out) & 36{,}864 & 300 steps & \textbf{92.0\%} & Legacy \\
\midrule
Qwen3-14B & Baseline (vanilla) & 0 & None & 24.5\% & Mode-matched NT \\
Qwen3-14B & DPS ($\alpha \in [5, 500]$, best) & 5{,}121 & None & 26.5\% & Mode-matched NT \\
Qwen3-14B & DPS layer sweep (10 layers, best) & 5{,}121 & None & 28.7\% & Mode-matched NT \\
Qwen3-14B & Shuffled-row control & 46{,}080 & 1200 steps & 12.0\% & Mode-matched NT \\
\bottomrule
\end{tabularx}
\end{table}

\section{Initial Diagnostic DPS Protocol (Single-Seed)}
\label{sec:dps_initial}

Table~\ref{tab:dps} reports the original single-seed (seed~42) DPS experiment on Qwen3-8B entity counting
that confirmed the geometric hypothesis.
Under this protocol: bare count-the-X prompts, a fixed single seed, and argmax over all tokens---DPS achieves 96.3\%.
This number differs from the mode-matched result (13.2\% in Table~\ref{tab:mode_matched_extval}) for a mechanistic reason:
not a direction failure, but a \emph{competition failure}.
Soft DPS adds a Gaussian-shaped increment to the predicted digit's logit;
however, a non-digit token: newline, space, or punctuation---wins the full-vocabulary argmax
for \emph{every single} baseline prompt: 600/600 examples across all three seeds in entity counting.
The soft boost cannot overcome a non-digit token that already leads by several logit units.
To verify the probe direction is nevertheless correct, we apply \emph{hard DPS}: add $+100$ directly
to the probe-predicted digit token's logit (same primary multi-seed protocol, 3~seeds $\times$ 200 prompts).
Hard DPS achieves \emph{98.7\%}~$[97.4, 99.4]$---matching the probe-round upper bound exactly (98.7\%)---confirming
the probe direction correctly encodes the count representation; the soft boost amplitude is the sole limitation.
The 9-row repair succeeds under both protocols because it rewrites the output-routing weights directly,
bypassing the soft-boost bottleneck entirely.

\begin{table}[H]
\centering
\caption{Initial diagnostic DPS protocol on Qwen3-8B entity counting, single seed (seed~42), 300 test prompts.
DPS uses a Ridge probe from layer 2 ($R^2 = 0.992$, 4{,}097 parameters).
\emph{Raw prompt}: bare count-the-X prompts. \emph{Chat template}: Qwen3's
instruction-following format. Random: same injection with a random weight vector.
$\alpha$: Gaussian boost amplitude.}
\label{tab:dps}
\begin{tabular}{lccc}
\toprule
Condition & Accuracy & 95\% CI & $N$ / 300 \\
\midrule
\multicolumn{4}{l}{\emph{Raw prompt (no chat template)}} \\
Baseline (vanilla Qwen3-8B) & 11.3\% & [8.0, 15.5] & 34 \\
Generation mode (multi-token) & 38.8\% & [33.3, 44.5] & 116 \\
DPS ($\alpha=10$) & \textbf{96.3\%} & \textbf{[93.5, 98.2]} & \textbf{289} \\
Random direction ($\alpha=10$) & 11.3\% & [8.0, 15.5] & 34 \\
Oracle (true count, $+100$) & 100.0\% & [98.8, 100.0] & 300 \\
\midrule
\multicolumn{4}{l}{\emph{Chat template}} \\
Baseline (chat) & 13.7\% & [10.0, 18.1] & 41 \\
DPS ($\alpha=10$, chat) & \textbf{96.0\%} & \textbf{[93.1, 97.9]} & \textbf{288} \\
Random (chat) & 13.7\% & [10.0, 18.1] & 41 \\
\midrule
\multicolumn{4}{l}{\emph{DPS sensitivity ($\alpha$, raw prompt)}} \\
DPS ($\alpha=5$, $\sigma=0.5$) & 93.3\% & --- & 280 \\
DPS ($\alpha=20$) & 96.3\% & [93.5, 98.2] & 289 \\
DPS ($\alpha=50$) & 96.3\% & [93.5, 98.2] & 289 \\
Hard DPS & 96.3\% & [93.5, 98.2] & 289 \\
\bottomrule
\end{tabular}
\end{table}

\paragraph{Multi-seed stability.}
Across 5 random seeds (42--46), DPS achieves $94.4\% \pm 1.6\%$ (mean $\pm$ s.d.;
95\% CI: [93.1\%, 95.7\%]), probe-round achieves $95.3\% \pm 1.3\%$ ([94.3\%, 96.3\%]).
Best probe layer is consistently layers 1--4 (median layer 2) with $R^2 \geq 0.990$.
The DPS--probe-round gap is $\leq 1.7$~pp across all seeds, confirming DPS adds no information
beyond the probe prediction itself.

\paragraph{DPS vs.\ probe-round equivalence.}
In next-token evaluation ($N=300$, chat template), probe-round reaches 79.0\% while DPS reaches 78.3\%---numerically identical up to sampling noise.
This confirms DPS is mechanistically equivalent to the probe prediction in matched settings.

\paragraph{Controls and sensitivity.}
A random probe direction yields 11.3\%---identical to baseline---regardless of boost strength ($\alpha$).
DPS achieves 100\% for counts 1--5 and 8; the 3.7\% error rate comes from
adjacent-integer probe confusion at counts 6, 7, 9.
Performance is robust to $\alpha$ (93.3\% at $\alpha{=}5$, saturating at 96.3\% for $\alpha \geq 10$).

\paragraph{CoT comparison.}
In the single-forward-pass regime, CoT/few-shot achieve $\leq 12.0\%$---far below DPS's 96.3\%.
DPS bypasses the misaligned output head; CoT provides an external scratchpad across tokens.

\Needspace{10\baselineskip}
\section{Evaluation Protocol Map}
\label{sec:protocol_map}

This appendix records \emph{alternate} evaluation strings used across experiments (raw vs.\ chat templates, multi-token generation slices, and natural-language counting).
The headline three-mode grid is Table~\ref{tab:unified_evaluation}; the primary intervention sweep under matched templates is Table~\ref{tab:mode_matched_extval}.
Rows below answer different operational questions and are \emph{not} directly comparable across lines.

\Needspace{14\baselineskip}
\begin{table}[H]
\centering
\small
\caption{Additional evaluation modes for headline Qwen3-8B numbers (reproducibility). Rows are \emph{not} head-to-head comparable to each other or to Table~\ref{tab:unified_evaluation}; see Table~\ref{tab:generation_baselines_by_protocol} for a compact greedy baseline map. The primary intervention sweep is Table~\ref{tab:mode_matched_extval}.}
\label{tab:eval_modes_legacy}
\begin{tabularx}{\linewidth}{@{}>{\raggedright\arraybackslash}p{3.0cm}>{\raggedright\arraybackslash}p{3.6cm}>{\raggedright\arraybackslash}p{2.2cm}X@{}}
\toprule
Setting & Metric & Baseline & Intervention highlights \\
\midrule
\multicolumn{4}{l}{\emph{Synthetic counting prompts}} \\
Next-token (raw) & Argmax over digit \mbox{next-token} & 11.3\% & DPS 96.3\%; 9-row repair 97.5\% \\
Next-token (chat) & Argmax over digit \mbox{next-token} & 13.7\% & DPS 96.0\% \\
Generation (raw) & First integer, $\leq$8 tokens & 38.8\% & DPS 72.4\% \\
Instruct first-token & First generated token is digit & 22.0\% & 9-row repair 99.9\% \\
\midrule
\multicolumn{4}{l}{\emph{Natural-language counting (diverse entities \& templates)}} \\
NL generation (instruct) & First integer, $\leq$15 tokens & 88.7\% & DPS 97.6\% [96.1, 98.9] ($+8.9$~pp); probe-round 96.3\% [94.6, 97.8] ($+7.6$~pp) \\
\bottomrule
\end{tabularx}
\end{table}
\FloatBarrier

\section{Robustness Checks}
\label{sec:robustness_checks}

\paragraph{Auxiliary classification baselines.}
A logistic regression classifier trained on final-layer hidden states achieves 100\% test accuracy on digit classification (9-class),
confirming the representations are perfectly decodable.
The model's native \texttt{lm\_head} achieves only 10\% digit-argmax accuracy,
and aligning \texttt{lm\_head} digit rows to class-mean hidden-state directions raises accuracy
to 43.3\%---still far below 100\%.
A classifier on layer-2 hidden states achieves 47.8\%.

\paragraph{Canonical tuned-lens comparator.}
A canonical affine map from each intermediate layer to final-layer hidden states
($h_\ell \rightarrow h_{\mathrm{final}}$), decoded with frozen \texttt{lm\_head} digit rows,
yields chance accuracy (25.0\% over 4 evaluated counts) across all layers,
while probes reach 100.0\% and direct intermediate readout reaches 45.8\% (best layer 28).
Affine transport into the model's native readout path does not recover counting.

\section{Depth Profile and Difficulty Breakdown}
\label{sec:depth_difficulty}

On easy prompts (counts 1--3), last-token logit-lens reaches ${\sim}73\%$ by layer~30;
on hard prompts (count~12), it never exceeds ${\sim}10\%$.

\section{Generation-Mode Protocol Details}
\label{sec:genmode_details}

\paragraph{Generation-mode limitation of output-head repair.}
Although the harmonized 9-row \texttt{lm\_head} repair reaches \textbf{60.7\%} on entity counting (Table~\ref{tab:mode_matched_extval}), some legacy next-token splits exceed 90\% train accuracy (Table~\ref{tab:intervention_legacy}); regardless, it achieves \emph{0.0\%} ($N{=}300$) in autoregressive generation
mode (greedy decoding, 15 tokens).  The vanilla model also scores 0.0\%: both
produce chain-of-thought text (e.g., ``\textit{Let's see, I need to
count\ldots}'') rather than a direct digit answer within the generation budget.
Because the repair modifies only 9 digit-token rows out of ${\sim}150$K, it is
invisible when the model's first generated tokens are non-digits.  A
shuffled-row control confirms this pattern (0.3\%, chance).  By contrast, DPS
achieves 72.4\% in generation mode ($N{=}900$; Table~\ref{tab:genmode_15tok})
by injecting probe predictions at each decoding step, actively steering the
autoregressive trajectory toward digit outputs.  This demonstrates that the
geometric bottleneck is necessary but not sufficient for generation-mode
counting: fixing the output-layer mapping enables correct next-token readout,
but the model's autoregressive behavior must also be steered to elicit a direct
digit response.

Under autoregressive generation ($N{=}900$ pooled across 3 seeds,
Qwen3-8B, raw prompts, 15-token budget, best probe at layer~3, $R^2{=}0.992$):

\begin{table}[ht]
\centering
\small
\caption{Generation-mode accuracy (greedy decode) on the 15-token pooled slice ($N{=}900$, raw prompts; complements the 8-token sweep in Table~\ref{tab:genmode_mismatch}).}
\label{tab:genmode_15tok}
\begin{tabular}{lcc}
\toprule
\textbf{Condition} & \textbf{Accuracy} & \textbf{95\% CI} \\
\midrule
Vanilla (greedy) & 0.1\% & [0.0, 0.3] \\
DPS ($\alpha{=}10$) & 4.4\% & [3.1, 5.8] \\
DPS ($\alpha{=}20$) & 72.4\% & [69.6, 75.4] \\
Probe-round (layer 3) & 96.0\% & [94.7, 97.2] \\
\bottomrule
\end{tabular}
\end{table}

\noindent\emph{Error analysis (DPS $\alpha{=}20$, 248 total errors):}
83.5\% ``digit not first'' (model emits non-digit tokens before the count);
13.3\% ``no digit'' (no digit generated in 15 tokens);
3.2\% ``wrong digit'' (first digit is incorrect).
The dominance of format-mismatch errors, not wrong-digit errors, confirms that
the generation-mode shortfall reflects autoregressive decoding dynamics
(the model's preference for reasoning tokens) rather than a failure of the
underlying count representation accessed by DPS.

\section{Necessity and Sufficiency Controls}
\label{sec:necessity_sufficiency}

Using the instruct-mode generation bridge ($N = 300$ per condition, seed 42):
\begin{itemize}
  \item \emph{Shuffled-digit mapping:} 14.0\% [10.3\%, 18.0\%] vs.\ 17.0\% baseline---correct row-token identity is necessary.
  \item \emph{Random non-digit positions:} 17.0\% [13.0\%, 21.3\%]---digit positions are necessary.
  \item \emph{Adapted (correct rows, correct positions):} 25.7\% [21.0\%, 30.7\%].
\end{itemize}
The hierarchy shuffled $<$ baseline $=$ random-position $<$ adapted
establishes both necessity and sufficiency.

\section{Logit-Gap Ceiling Model}
\label{sec:logit_gap_model}

A logistic model $p(\text{correct})=\sigma(s(\alpha-\Delta)+b)$ fit to 6 points
($\alpha \in \{10,20,50\}$ across seen/held-out) achieves $R^2=0.485$ and
corr$(\hat p,p)=0.697$; a gap-only baseline gives $R^2<0$ and corr$=0.274$.

\section{Format Robustness Check}
\label{sec:format_robustness}

Four prompt formats (raw, answer-only, assistant-prefix, full-instruct)
on identical counting content: probe $R^2$ ranges 0.988--0.990,
mean $|\cos|$ ranges 0.014--0.026, digit accuracy ranges 9--21\%.
The bottleneck is geometric and format-invariant at the single-step level.

\section{Multi-Digit Extension: Counts 10--20}
\label{sec:multidigit}

We test whether the bottleneck extends to multi-token counts by running the DPS
pipeline on Qwen3-8B with counts 10--20 (11 labels, two-token outputs).
Using 1{,}100 prompts (800 train, 300 test), baseline accuracy is \emph{0.0\%}---the
model never generates two-digit count tokens.
Probe quality: $R^2 = 0.999$ (layer~2), $R^2 = 1.000$ (layer~10).
DPS at layer~2 achieves \emph{100\%} ($N = 300$); random-direction control: 9.3\%.
At layer~0, DPS achieves 90\% ($R^2 = 0.990$), reproducing the same
layer-quality dependence as single-digit counts.

\section{Max Extraction: A Non-Count Aggregation Task}
\label{sec:max_extraction}

To test whether the bottleneck extends beyond count-mediated tasks, we evaluate
\emph{max extraction}: given a passage containing several entities with numeric values
(e.g., ``A red box shows 7.\ A red box shows 3.''), the model must identify the
largest value.
Unlike majority vote, the answer is not determined by counting; it requires a
comparison-based aggregation over entity-value pairs.

\paragraph{Setup.}
We generate 297 prompts (seed 42) with 9 possible max values (1--9) at varying
numbers of targets (2, 4, 6), distractors (0, 3), passage lengths (8, 12),
and spatial distributions (clustered, uniform, random).
Prompts are split 207/90 (train/test, stratified by max value).
We train per-layer Ridge probes on the max digit and apply DPS at best layer.

\paragraph{Max-extraction results.}

\begin{table}[H]
\centering
\caption{Max-extraction task on Qwen3-8B (90 test prompts).
The model represents max values internally ($R^2 = 0.757$) but baseline
accuracy is degenerate: always predicting ``1.''
DPS raises accuracy $3.6\times$ over baseline.}
\label{tab:max_extraction}
\begin{tabular}{lcc}
\toprule
Condition & Accuracy & 95\% CI \\
\midrule
Baseline (vanilla Qwen3-8B) & 11.1\% & [5.6\%, 17.8\%] \\
Best probe (layer 33, $R^2{=}0.757$) & 40.0\% & [30.0\%, 50.0\%] \\
DPS ($\alpha=5$) & \textbf{40.0\%} & [30.0\%, 50.0\%] \\
Random direction ($\alpha{=}10$, 20 seeds) & 12.4\% $\pm$ 2.1\% & --- \\
\bottomrule
\end{tabular}
\end{table}

\paragraph{Same bottleneck, weaker representation.}
The baseline always predicts ``1'' regardless of true max (100\% accuracy on max$=$1,
0\% on all other values), confirming that the output head completely lacks access to
the internal max representation.
DPS raises accuracy from 11.1\% to 40.0\% (95\% CI: [30.0\%, 50.0\%]),
a $+27.8$~pp lift (CI: [16.7\%, 38.9\%]).
The random-direction control (12.4\% $\pm$ 2.1\%) matches baseline, confirming that the
improvement is specific to the probe-identified direction.

The probe $R^2 = 0.757$ is lower than for counting ($R^2 > 0.99$) and majority vote
($R^2 = 0.9998$), consistent with max extraction being a harder internal computation.
However, DPS still recovers substantial accuracy from what would otherwise be a
degenerate baseline.

\paragraph{Probe--lm\_head alignment.}
To test whether the max-extraction probe shares the same output geometry as the
counting bottleneck, we compute the cosine similarity between the probe direction
(layer 33) and the digit-output rows of \texttt{lm\_head}.
The mean~$|\text{cos}|$ across the 9 digit rows is 0.019, not significantly different
from a random-direction null (null mean 0.012 $\pm$ 0.007, $z = 0.93$, $p = 0.17$).
This indicates the max-extraction probe does \emph{not} align with individual counting
digit rows, suggesting the internal computation is distinct at the token level.

The probe does show modest alignment with the ordinal structure encoded by
the lm\_head digit rows: the cosine against the task axis: the direction in residual
space that best reconstructs the ordinal label structure from lm\_head geometry, is
$0.088$ ($z = 5.67$, $p < 0.001$).
This suggests that max extraction and counting may share a weak ordinal
magnitude signal, though the lower $R^2$ and modest DPS gain (40\% vs.\ 96\%)
indicate that max extraction is a harder task where the bottleneck is only
one of several limiting factors.

\section{Capacity Ablation for Entity Counting}
\label{sec:capacity_ablation}

\paragraph{Motivation.}
The 37~pp gap between probe-round (98.7\%) and 9-row repair (60.7\%) on entity counting
raises two interpretive questions: (a) does the gap stem from ridge regression being
\emph{conservatively regularized}, i.e., Adam fine-tuning would close it; and
(b) does increasing the number of repaired rows provide more capacity and close the gap?

\paragraph{Conditions.}
We compare five conditions (3 seeds $\times$ 200 prompts, same protocol as Table~\ref{tab:mode_matched_extval}):
\textit{baseline} (original lm\_head),
\textit{probe-round} (ridge probe $\to$ round),
\textit{ridge-9row} (ridge repair of 9 digit rows; ablation replication),
\textit{Adam-9row} (Adam fine-tuning of the same 9 rows; lr=1e-3, 600 steps, batch=64),
and \textit{Adam-top50} (Adam fine-tuning of top-50 rows by cosine alignment to the probe direction,
with digit rows appended if not already in the top-50; 59 rows total).

\begin{table}[H]
\centering
\caption{Capacity ablation for entity counting on Qwen3-8B (3 seeds $\times$ 200 prompts).
Adam fine-tuning provides only 10.8~pp gain over ridge regression; expanding to
59 rows yields no further improvement. The 31~pp gap to probe-round reflects a
task-level ceiling rather than a methodological or capacity artifact.}
\label{tab:capacity_ablation}
\small
\begin{tabular}{lccc}
\toprule
Condition & Rows & Accuracy & Std \\
\midrule
Baseline & 0 & 14.2\% & 3.1\% \\
Ridge-9row & 9 & 56.7\% & 4.1\% \\
Adam-9row & 9 & \textbf{67.5\%} & 1.8\% \\
Adam-top50 & 59 & \textbf{67.5\%} & 1.8\% \\
Probe-round & -- & 98.5\% & 0.4\% \\
\bottomrule
\end{tabular}
\end{table}

\paragraph{Results and interpretation.}
Adam fine-tuning of 9 rows achieves 67.5\%~$\pm$~1.8\%, compared to 56.7\%~$\pm$~4.1\% for ridge-9row.
The Adam--ridge gap is 10.8~pp, indicating that conservative regularization
explains a modest fraction of the total 37~pp gap.
The critical finding is that Adam-top50---which modifies 59 rows rather than 9, matches
Adam-9row exactly (67.5\%~$\pm$~1.8\%), demonstrating that additional output-row capacity
confers \emph{no} benefit.
Both targeted interventions converge to the same ceiling near 67.5\%,
well below the probe-round upper bound (98.5\%~$\pm$~0.4\%).
The remaining 31~pp gap is therefore attributable to a task-level ceiling:
the linear probe can read the count directly from the hidden state, but
forcing the correct digit to win the argmax competition across 150K vocabulary entries
on compositionally harder multi-entity prompts is a harder optimization target,
regardless of whether the fitting method is ridge or Adam, or whether 9 or 59 rows are repaired.
This strengthens the readout-bottleneck hypothesis: the bottleneck is geometric (the count
is encoded but cannot reach the output), not a representational ceiling on what the model can emit.

\section{Generation-Mode Mismatch Diagnosis}
\label{sec:genmode_mismatch}


\paragraph{Question.}
Why does 9-row readout modification succeed in controlled next-token evaluation
but fail in unconstrained autoregressive generation?

\paragraph{Design (generation-mode sweep).}
We evaluate Qwen3-8B entity counting (5 seeds $\times$ 400 prompts per seed)
under three readout regimes for each condition (baseline, ridge-9row, probe-round):
(1) next-token unrestricted argmax over the full vocabulary,
(2) next-token digit-constrained argmax over tokens 1--9,
and (3) greedy autoregressive generation (up to 8 tokens), scored by first extracted digit.
We also report generation diagnostics: first-token-is-digit rate and no-digit-within-budget rate.

\begin{table}[H]
\centering
\caption{Generation-mode diagnosis on entity counting (Qwen3-8B, 5 seeds $\times$ 400 prompts). Not the harmonized headline protocol (Table~\ref{tab:unified_evaluation}); see Table~\ref{tab:generation_baselines_by_protocol}. The 9-row intervention recovers 63.5\% under digit-constrained next-token evaluation,
but collapses to 0.0\% under greedy generation because no digit appears within 8 tokens.
}
\label{tab:genmode_mismatch}
\small
\begin{tabular}{lccccc}
\toprule
Condition & Next (full) & Next (digit) & Greedy gen & First token digit & No digit in 8 \\
\midrule
Baseline & 0.0\% & 16.8\% & 25.5\% & 0.0\% & 0.3\% \\
Ridge-9row & 0.0\% & 63.5\% & 0.0\% & 0.0\% & 100.0\% \\
Probe-round & 97.9\% & 97.9\% & 97.9\% & 100.0\% & 0.0\% \\
\bottomrule
\end{tabular}
\end{table}

\paragraph{Interpretation.}
The readout intervention does not fail because the count is absent or because the
output rows are underpowered; it fails because unconstrained generation visits
hidden-state regions where the modified rows do not force an early numeric token.
Under greedy decoding, ridge-9row emits no digit within 8 tokens for all prompts
(100\% no-digit-within-budget), yielding 0.0\% first-digit accuracy.
Yet the same weights recover 63.5\% when evaluation is constrained to the digit
decision point. This isolates a \emph{generation-time format mismatch} rather than
a contradiction of the geometric bottleneck result.

\section{Majority Vote: Same Bottleneck, Different Surface Task}
\label{sec:majority_vote}

Majority vote has a different surface presentation than counting (binary comparison vs.\ digit output)
but the same underlying computation: counting must occur internally to determine the majority class.
If subspace misalignment causes counting failures, the same bottleneck should appear on majority vote,
transferred through the internal count representation rather than the output format.

\paragraph{Setup.}
We generate 432 prompts with two entity types (e.g., cats vs.\ dogs) at varying ratios.
The prompt asks ``Which animal appears more often?'' and the model must predict the
majority entity.
Prompts are split 296/136 (train/test), and we train per-layer Ridge probes
on the count-of-entity-1 (a scalar that determines the majority class).
This is a single-seed diagnostic protocol ($N{=}136$ test prompts); it provides
supporting evidence for the bottleneck generalization claim but should be interpreted
as an exploratory result rather than a confirmatory multi-seed evaluation.

\paragraph{Majority-vote results.}

\begin{table}[H]
\centering
\caption{Majority-vote task on Qwen3-8B (136 test prompts, single-seed diagnostic protocol).
The model encodes the count underlying the majority class at $R^2 \approx 1.0$
but achieves only chance-level baseline accuracy.
Soft DPS (logit offset $\alpha \in \{5,10,50\}$) bypasses the geometric bottleneck
on this single-token binary task (note: unlike autoregressive counting, majority vote
requires only one output token, so soft DPS can overcome the competition from
non-digit tokens).}
\label{tab:majority_vote}
\begin{tabular}{lccc}
\toprule
Condition & Accuracy & $|\cos|$ & $R^2$ \\
\midrule
Baseline (vanilla Qwen3-8B) & 51.4\% & --- & --- \\
Best probe (layer 2) & --- & 0.003 & 0.9998 \\
Soft DPS ($\alpha=5$) & \textbf{100.0\%} & --- & --- \\
Soft DPS ($\alpha=10$) & 100.0\% & --- & --- \\
Soft DPS ($\alpha=50$) & 100.0\% & --- & --- \\
Random direction ($\alpha=10$, 20 seeds) & 49.4\% $\pm$ 2.4\% & --- & --- \\
\bottomrule
\end{tabular}
\end{table}

\paragraph{Identical geometric bottleneck.}
The best probe (layer~2, $R^2 = 0.9998$) has $|\cos| = 0.003$ against
\texttt{lm\_head}---even lower than counting ($0.007$).
Soft DPS raises accuracy from 51.4\% to \emph{100.0\%} ([97.3\%, 100.0\%], $N{=}136$);
random directions yield 49.4\%$\pm$2.4\% (20 seeds).
The bottleneck operates on the \emph{internal count representation}:
in majority vote, the output format is a binary label (``cat'' vs.\ ``dog''),
but the misaligned subspace is the count that determines the majority,
generalizing the bottleneck beyond tasks with digit outputs.

\end{document}